\documentclass{article}
\usepackage{iclr2025_conference,times}

\usepackage{hyperref}
\usepackage{url}

\usepackage{algorithm}
\usepackage{algpseudocode}
\usepackage{amsmath}
\usepackage{cleveref}

\usepackage{booktabs}
\usepackage[breakable,skins,many]{tcolorbox}
\usepackage{tablefootnote}
\usepackage{setspace}
\usepackage{tikz}
\usepackage{jigsaw}

\usepackage{booktabs}
\usepackage{wrapfig}

\newcommand{\Agentsroom}{\textsc{Agents' Room}}
\newcommand{\Agentsroomzs}{\textsc{Agents' Room$_{ZS}$}}

\newcommand{\agentsroomzs}{{\sc AR$_{ZS}$}}
\newcommand{\agentsroomft}{{\sc AR$_{FT}$}}

\newcommand{\baseline}{{\sc E2E}}
\newcommand{\baselinezs}{{\sc E2E$_{ZS}$}}
\newcommand{\baselineft}{{\sc E2E$_{FT}$}}

\newcommand{\tellmeastory}{{\sc Tell$\,$me$\,$a$\,$story}}

\definecolor{my-green}{HTML}{72d0b3}
\definecolor{my-blue}{HTML}{99bcdc}
\definecolor{my-red}{HTML}{f19ed5}
\definecolor{my-purple}{HTML}{c5add4}
\definecolor{my-yellow}{HTML}{faea93}

\title{Agents' Room:  Narrative Generation through Multi-step Collaboration}

\author{\vspace{0.08cm}Fantine Huot,\, Reinald Kim Amplayo,\, Jennimaria Palomaki,\, Alice Shoshana Jakobovits, \\
\vspace{0.08cm}
\bf{Elizabeth Clark\, \&\, Mirella Lapata} \\
\vspace{0.08cm}
Google DeepMind \\
\small{\texttt{\{fantinehuot,reinald,jpalomaki,jakobovits,eaclark,lapata\}@google.com}}
}

\iclrfinalcopy % Uncomment for camera-ready version, but NOT for submission.

\begin{document}
\maketitle

\begin{abstract}
Writing compelling fiction is a multifaceted process combining
elements such as crafting a plot, developing interesting characters,
and using evocative language. While large language models (LLMs) show
promise for story writing, they currently rely heavily on intricate
prompting, which limits their use. We propose \Agentsroom, a
generation framework inspired by narrative theory, that decomposes
narrative writing into subtasks tackled by specialized agents. To
illustrate our method, we introduce \tellmeastory\footnote{We release
the dataset and metrics at: \small{\url{https://github.com/google-deepmind/tell_me_a_story}}}, a high-quality dataset of complex writing prompts and human-written stories, and a novel  evaluation framework designed specifically for assessing long narratives. We show that \Agentsroom\ generates stories that are preferred by expert evaluators over those produced by baseline systems by  leveraging  collaboration and specialization to decompose the complex story writing task into tractable components. We provide extensive analysis with automated and human-based metrics of the generated output. 
\end{abstract}

\section{Introduction}

% \mirella{In charge of this section} 

Creating long-form content requires meticulous research, advanced
planning, an engaging writing style, and the ability to craft stories
that captivate.  J.K. Rowling is claimed to have had most of the Harry
Potter story planned out before she started writing.  She knew there
would be seven books, which characters would be important and how they
would develop, and which key plot twists would serve the overall
story. In addition, she carried out substantial research to create the
fictional universe which provides the backdrop of the story. Breaking
down a story into distinct sections is typical for longer narratives,
with most stories boiling down to a few shared elements like
exposition, rising action, climax, falling action, and resolution
\citep{freytag1896freytag,pavis1998dictionary}. Practical guides to
writing successful screenplays \citep{cutting2016narrative,Hauge:2017}
outline a similar structure, following the setup, the new situation,
progress, complications and higher stakes, the final push, and the
aftermath.

%Academic papers \citep{schimel:2012} follow a common
%format which can be distilled into title, abstract, introduction, main
%body, conclusions, and references, with well-known conventions about
%the contents of each section (e.g., the introduction should convey the
%essence of the author's argument or idea).  Writing news articles
%\citep{ricketson:graham:2017} also involves standard subtasks like a
%research phase (gathering material), an outline phase (planning the key points in each section), and a writing phase.

\begin{figure}[t]
\centering
% trim={left bottom right top}
\includegraphics[width=0.7\linewidth, trim={4cm 12cm 4cm 3.7cm}, clip]{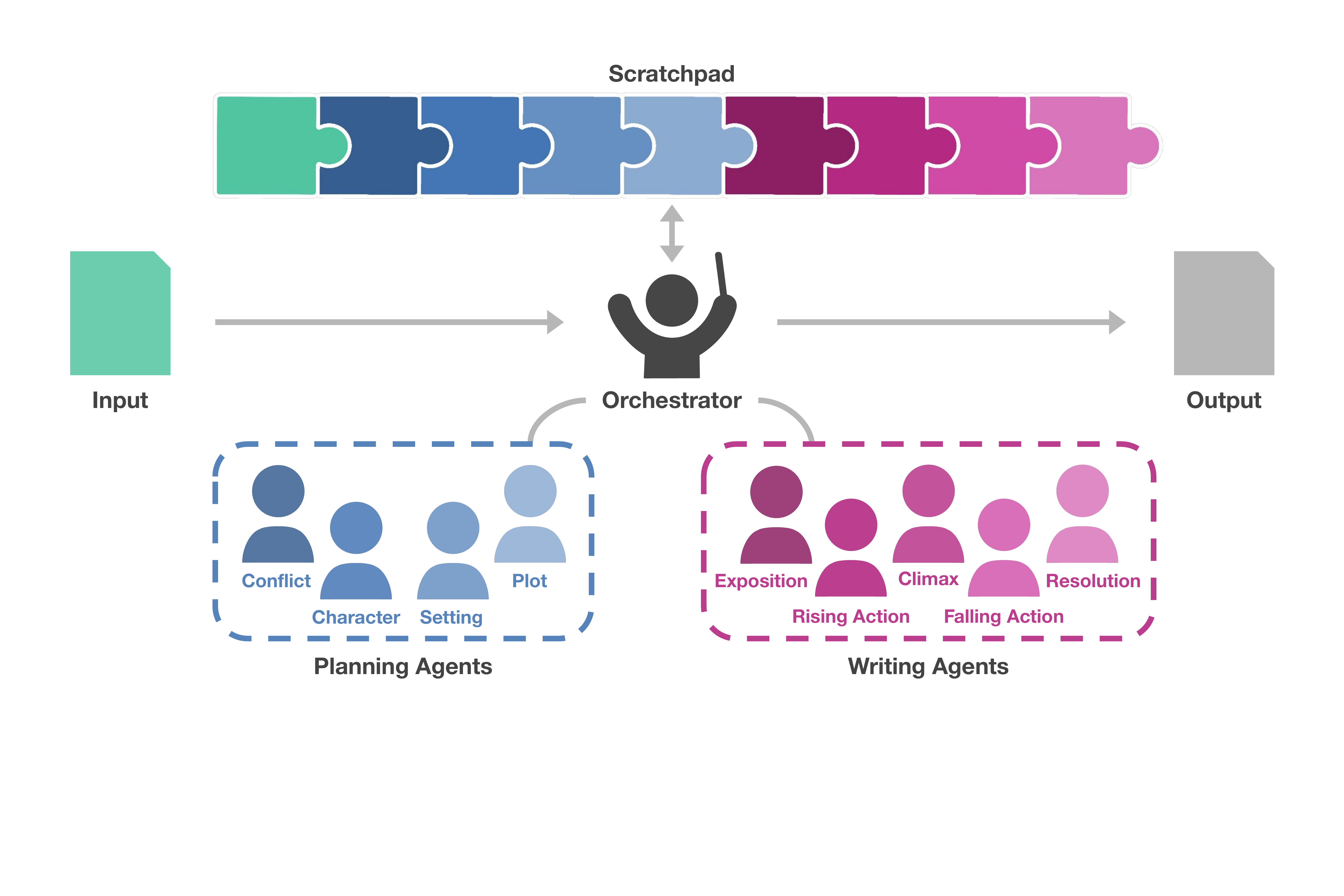}
\vspace{-.2cm}
\caption{\label{fig:agents_room} \Agentsroom, a multi-agent framework for collaborative writing. A central orchestrator calls the individual agents and consolidates their contributions into the scratchpad. We color-code each piece of the scratchpad with the contributing agent's color.}
\end{figure}

Large language models (LLMs) have demonstrated impressive writing
capabilities
\citep{yang-etal-2022-re3,fitria:2023,shao-etal-2024-assisting,bai2024longwriterunleashing10000word},
however, generating long-form content is still a challenge. Well-known
problems include maintaining a consistent narrative, tone, or factual
accuracy over extended stretches of text
\citep{chakrabarty:ea:2024b,WANG2023126792,alabdulkarim-etal-2021-automatic,balepur-etal-2023-expository,yamshchikov-tikhonov-2023-wrong},
and showcasing a unique voice or humor that makes writing truly
memorable. Despite displaying flashes of creativity, they often
replicate patterns found in their training data, which hinders the
generation of original concepts, plotlines, or phrasing. Added
problems include the lack of datasets or benchmarks for long-form
writing \citep{bai2024longwriterunleashing10000word} and standardized evaluation criteria for assessing creative writing
either by humans or machines
\citep{chhun-etal-2022-human,chhun2024languagemodelsenjoystories,chakrabarty:ea:2024a}.

Existing methods often rely on detailed prompts to guide the
generation process \citep{yang-etal-2022-re3,xie-etal-2023-next},
prompt chaining \citep{piotr:ea:2023,yang-etal-2022-re3}, and planning
strategies
\citep{yang-etal-2023-doc,lee2024navigatingpathwritingoutlineguided}
as a means of breaking down the complex writing task into more
manageable components. In this paper, we conceptualize long-form writing as a multi-agent collaboration problem. Rather than attempting
a decomposition of the writing task within a \emph{single} agent
\citep{chen2023program,yao:ea:2024}, we leverage collaboration among
\emph{multiple} agents, with specialized abilities
\citep{talebirad2023multiagentcollaborationharnessingpower,zhang-etal-2024-exploring,han2024llmmultiagentsystemschallenges}. We
propose \Agentsroom\footnote{\Agentsroom\ is  very loosely modeled after writers' room,  a collaborative space where writers, (usually of a television series), come together to write and refine scripts.} (\Cref{fig:agents_room}), a generation paradigm which consists of two types
of agents, namely \emph{planning} and \emph{writing} agents. Planning
agents flesh out key components of the content but do not write the
story as such. For example, a planning agent might specialize in
character descriptions, whereas another might focus on the plot or
central conflict. Writing agents are responsible for generating the
final output text and are also specialized, e.g., one may focus on the
introduction, and another on the conclusions.  The two types of agents
work collaboratively to complete the writing task, sharing and
managing information through a \emph{scratchpad} which maintains
outputs from planning agents and makes them available to writing
agents. An \emph{orchestrator} is responsible for calling the agents in
order depending on the task at hand.

Compared to single LLM-powered agents, this multi-agent approach offers several advantages:

\begin{itemize}
\item LLMs can be  specialized into various distinct agents (e.g., zero-shot prompted or fine-tuned) performing a single function  with high precision; 
\item it avoids well-known problems with lengthy and under-specified
  instructions which require multiple iterations to build context and
  fully define an appropriate solution;
\item it can be applied to problems whose solution is not known beforehand, and results from exploring a vast research space or involves very long output (e.g., writing  a book); 
\item it naturally lends itself to human-in-the loop automation where
  machine-based agents can be replaced with human ones when needed.
\end{itemize}

We formalize \Agentsroom\ as a general writing framework and apply it
to creative writing. Specifically, we focus on writing long-form
stories (1,000-2,000 tokens), and create specialized agents drawing
inspiration from narrative theory (e.g., \citealt{card:1999,noble:1999,pavis1998dictionary}). To evaluate our
method, we introduce \tellmeastory, a new dataset of human-created writing
prompts and fiction stories, and a novel evaluation framework designed
for assessing multiple dimensions of story quality. Experimental
results show that \Agentsroom\ generates stories that are preferred (by
humans and automatic metrics) over those produced by baseline systems
which do not leverage collaboration or specialization.

\section{Related Work}
\label{sec:related_work}

% \mirella{In charge of this section}

\textbf{Story Generation} The advent of large pre-trained language
models has provided a common framework for generating stories which
sound fluent but often struggle with maintaining coherence and
plausibility.  Attempts to enhance coherence and control the
trajectory of events often decompose the generation task into planning
an outline or sketch, and then elaborating on it, e.g.,~by filling in
descriptions and specific details of each story. Examples of
intermediate plans include sequences of entities and their actions
\citep{yao2019plan}, outlines
\citep{fan-etal-2019-strategies,zhou2023recurrentgptinteractivegenerationarbitrarily,wang-etal-2023-improving-pacing},
plot structures \citep{goldfarb-tarrant-etal-2020-content}, and more
elaborate descriptions including details about the setting of the
story, its characters, and main plot points
\citep{yang-etal-2022-re3,yang-etal-2023-doc}. Other work uses common
sense knowledge to impose constraints on the characters and their
interactions \citep{peng-etal-2022-inferring}, ensemble-based models
to render event sequences more plausible
\citep{Ammanabrolu_Tien_Cheung_Luo_Ma_Martin_Riedl_2020}, 
stylistic constraints \citep{kong-etal-2021-stylized}, and 
twists through constrained decoding \citep{huang-etal-2023-affective}.
These efforts have demonstrated that generating stories as a one-step
process is challenging, and ultimately various interventions are
required to improve overall story quality. Our work follows on from
this realization, and breaks down the writing task into subtasks,
undertaken by different agents who collaboratively plan and write a
story. Collaborative writing is often used in academic or professional contexts to leverage the strengths
and perspectives of various contributors, and has also been shown to
enhance creativity \citep{barrett:ea:2021}. 

Using LLMs as tools to assist humans with writing stories is an active
research area
\citep{chakrabarty:ea:2024b,piotr:ea:2023,ippolito2022creativewritingaipoweredwriting}. In
our experiments, stories are written exclusively by models without
humans in the loop.  However, our framework is fairly general allowing
for human-machine collaboration at various stages of content creation.

\textbf{Multi-agent Systems}
LLM-based agents have recently shown robust reasoning and planning
capabilities across various application domains
\citep{zhao2023surveylargelanguagemodels,bubeck2023sparksartificialgeneralintelligence}. Multi-agent
systems involve multiple independent LLMs working together to solve
complex tasks that are beyond the capability of any individual agent
\citep{talebirad2023multiagentcollaborationharnessingpower,park:ea:2023,han2024llmmultiagentsystemschallenges,ijcai2024p890}.
The agents are typically specialized in different aspects of a problem
or have different roles, allowing the system to approach tasks in a more coordinated, distributed, and modular way.  LLM-based multi-agent
systems have recently demonstrated promising results in multiple areas
including software development \citep{hong2024metagpt}, robotic tasks
such as motion planning \citep{mandi:ea:2024}, simulations of human
behavior \citep{park:ea:2023,hua2024warpeacewaragentlarge}, the
creation of gaming enviroments \citep{hu2024survey}, recommender
systems \citep{zhang:ea:2024},  simulations of financial trading
\citep{RePEc:arx:papers:2309.03736}, and policy making 
\citep{xiao2023simulatingpublicadministrationcrisis}.  We are not
aware of existing multi-agent frameworks for long-form writing. We
draw inspiration from related work demonstrating that collaborative
problem-solving improves LLM task-solving capabilities
\citep{hao2023chatllmnetworkbrainsintelligence,wang-etal-2024-unleashing,zhang-etal-2024-exploring}. Our
agents each adopt a specialized writing subtask and communicate
through a shared scratchpad (or memory) which allows to effectively
recall and utilize contextually-relevant past knowledge. In our
experiments, we predefine the number and type of agents best suited to
our story writing task, rather than dynamically generate agents
based on story content \citep{ijcai2024p3}.

\textbf{Evaluation}
Story evaluation is admittedly a challenging task for humans and
machines. Human evaluation is usually considered as the gold standard,
but it is expensive, time-consuming \citep{guan-huang-2020-union}, and
can be subjective. It also cannot capture diversity since a model that copies directly from the training set would potentially pass the human quality bar without displaying any generalization or creativity
\citep{hashimoto-etal-2019-unifying}. Automated evaluation metrics based
on lexical overlap or semantic similarity between generated stories and their human references have been shown to correlate poorly with
human judgements \citep{chhun-etal-2022-human}. In this paper, we
introduce an LLM-based evaluator
\citep{liusie2023zero,liu2024aligning,zheng2024judging,bohnet2024long}
to perform side-by-side comparisons of system outputs which correlates
with human judgements. Inspired by recent proposals on how to assess
human creativity \citep{chakrabarty:ea:2024a}, we distill the story
evaluation task into a few dimensions (e.g., plot, language use) which humans and machines can judge reliably.

\section{Agents' Room}

% reinald's comments
% formalize the multi-agent collaborative writing problem

% given agents A={a,...}
%  each a has a specific skill (e.g., write the exposition of the story), accessible as a function that accepts text inputs and returns 
%  one can be a planning agent or a writing agent. a planning agent 

% given a set of agents A, each with different skills F_a (e.g., writing a specific section) 
% 
% can be split into planning and writing agents
% planning agents are those that 

% \reinald{Create a figure for Agents' Room}

\begin{algorithm}[t]
\caption{\Agentsroom\ framework}
\label{alg:agents_room}
\begin{algorithmic}
\State $s \gets x$ \Comment Initialize the scratchpad
\While{$o(s, \mathcal{A})$ == True and $t < T$} \Comment While the orchestrator assigns a next agent
    \State $a_t = o(s, \mathcal{A})$ \Comment Select an agent given scratchpad
    \State $y_t = a_t(s)$ \Comment  Obtain agent's output
    \State $s \gets (s; (l_t, y_t))$ \Comment Update scratchpad
    \If{\text{type}($a_t$) == \text{writing}} \Comment If the agent is a writing agent, write to the final output
    \State $y \gets (y; y_t)$
    \EndIf
\EndWhile \\
\Return $y$ \Comment Return the final output
\end{algorithmic}
\end{algorithm}

% \fantine{In charge of this section}
In this section, we 
formalize \Agentsroom, the proposed multi-agent framework for collaborative writing. Given a complex writing task~$x$,  we generate output~$y$, by decomposing the writing process into multiple subtasks tackled by specialized agents. The full \Agentsroom\ framework is summarized in \Cref{alg:agents_room} and explained below.

\textbf{Agents} We define an {\it agent} $a \in \mathcal{A}$ as a
specialized model that takes text as input and returns text as output,
specified by a unique identifier label $l$ and a mapping $f:
\mathcal{V}^* \to \mathcal{V}^*$ (where~$\mathcal{V}$ are vocabulary
tokens). Each agent is specialized in a specific subtask. Under this
definition, an agent can be a LLM fine-tuned for the subtask, a
zero-shot prompted LLM with a specific input prompt, a deterministic
text processing function (e.g.,~string formatting and parsing),
% a tool call (e.g.,~calling a search engine, using a calculator), 
or even a human interacting with the system. Herein, we focus on
LLM-based agents, but we formalize the general framework's modeling
assumptions (e.g.,~agent inputs and outputs as text instead of latent
variables) to allow future work to incorporate human agents as well
(e.g., by iteratively editing LLM-generated text). We define two types
of agents (see below), namely {\it planning} and {\it writing} agents,
which differ both in function and in their mode of interaction with
the generated output.

% \fantine{Comment on the advantages of multi-agent systems: plug-n-play, individual components are easier to quality control, distributed system have engineering advantages, interfacing with privacy models, interfacing with proprietary models,  combining models of different sizes etc}

\textbf{Multi-agent Communication} Communication between agents is critical for the successful completion of their tasks. 
% There are at least three forms of communication: Collaborative, Debate, and Competitive. Collaborative agents work together towards a shared goal, exchanging information to enhance a collective solution. The Debate paradigm is employed when agents engage in argumentative interactions, presenting and defending their viewpoints or solutions and critiquing those of others. Competitive agents work towards their own goals that might conflict with other agents' goals. Herein, we focus on the {\it collaborative} communication framework since it is the one that would transfer most naturally to human-LLM collaborations, but the other alternatives are also possible. 
While different forms of communication are possible, such as debate \citep{khan2024debating,zhang-etal-2024-exploring} or competition \citep{cheng2024self}, in this work we focus on  {\it collaborative} communication  since it would transfer most naturally to human-LLM collaborations. Collaborative agents work together towards a shared goal, exchanging information to enhance a collective solution.

\textbf{Scratchpad} 
% Since the agents must work collaboratively to complete the writing task, 
The overall system requires a mechanism for sharing and managing
information across the different agents.  We assume our agents have
access to a shared {\it scratchpad} $s \in \mathcal{V}^*$ that
maintains individual agents' outputs and is passed along to the next
agent. The scratchpad is initialized with the initial writing
prompt~$x$ and is then updated after each agent call. At each
step~$t$, the current agent~$a_t$ takes as input the current
scratchpad $s_t$ and generates output~$y_t$. At each step, the
scratchpad is updated with the agent's unique identifier and output
such that $s_{t+1} \gets (s_t; (l_t, y_t))$. We include the agent's
label so that individual agents can easily reference and parse
specific portions of the scratchpad to complete their subtask.  Note
that in this framework, the scratchpad does not contain the specific
input prompt of a given LLM agent. Indeed, it is considered part of
each agent's subtask to process the output $y_t$ into a suitable
format to be used by other agents. Since agents have access to the
scratchpad, this means that they can avoid writing redundant and
duplicate information.

\textbf{Orchestrator} We have opted for a centralized architecture, where a central {\it orchestrator} determines the order upon which individual agents are called, and whether calling on each agent is necessary (e.g., depending on the task). Given a scratchpad $s_t$ and a set of available agents $\mathcal{A}$, the orchestrator $o: \mathcal{V}^* \times \mathcal{A}^* \to \mathcal{A}$ determines the next agent $a_{t+1}$ to call. It can be modeled as a Markov process, since each step depends entirely on the state of the scratchpad at step $t$. This orchestrator can be a discrete deterministic process, can have learnt transition probabilities, or can be arbitrarily complex. The orchestrator determines the stopping condition, i.e,~when there is no more agent to call, or when a maximum number of steps~$T$ has been reached.

\textbf{Planning Agents} Previous work (see \Cref{sec:related_work}) shows that LLMs benefit from an intermediate planning stage before generating the final output. 
% Concretely, this intermediate planning stage can be a chain-of-thought sequence \citep{wei2022chain}, a sequence of results from tool use \citep{schick2024toolformer}, a rewritten prompt with additional detail \citep{srivastava2023instance}, or a plan or outline of what to say and in what order \citep{narayan2023conditional}. 
These intermediate steps improve the overall output but are not included in the final output. We define {\it planning agents} as agents that specialize in generating these intermediate steps and write exclusively to the scratchpad.  For instance, when writing a story,  planning agents can draft character descriptions and plot elements; when writing an essay, they can outline the argumentative structure and retrieve references to substantiate claims. 
Since their outputs are in text format, a human-in-the-loop could review or edit these intermediate stages to guide the generative process.

\textbf{Writing Agents} Certain complex tasks, such as generating particularly long outputs or with sections written in different styles, remain challenging for a LLM to generate in one go. In such cases, the final output benefits from being generated section by section through separate agent calls. We define {\it writing agents} as agents specializing in writing specific parts of the final output. In addition to writing to the scratchpad, these writing agents iteratively write pieces of the final output $y$. Therefore, the final output can be formalized as the concatenation of the outputs of all the writing agents. For story writing, writing agents can specialize in specific parts of the narrative arc, such  as the exposition or the climax; when writing an essay, they can each tackle different sections, such as the arguments in favor versus against.

\section{Fiction Writing Task}

% Writing compelling fiction is a multifaceted process combining elements such as crafting a plot, developing compelling characters, and using evocative language. 
In this section, we present an instantiation of the
\Agentsroom\ framework for fiction writing: given an initial writing
prompt~$x$, generate narrative~$y$. We also introduce \tellmeastory \footnotemark[1], a
new high-quality dataset of human-created writing prompts and fiction
stories.

%\mirella{Create a figure for the fiction writing task}

%\vspace{-.2cm}

%\noindent\begin{minipage}{\textwidth}
%\captionof{figure}{Example prompts from the \tellmeastory\ dataset. %The corresponding stories are in %\Cref{appendix:fictionstories}.}\label{fig:prompts}
%\end{minipage}

% \jenni{Elaborate on the dataset creation}
\subsection{Specialized Agents Inspired by Narrative Theory}
\label{sec:fiction_agents}

% \fantine{In charge of this section}

% \mirella{Change examples to harry potter to match introduction}

We design specialized  agents for our fiction writing task by drawing inspiration from narrative theory. 
% Human writers usually draft the main story elements before crafting the final story. Similarly, 
We design four \emph{planning agents}, each specialized in a specific aspect of story planning: {\sc [conflict]} defines the central conflict (e.g.,~a young boy has to fight an evil wizard who killed his parents), {\sc [character]} develops the character(s) (e.g., the young man is brave, caring, determined, loyal to his friends with a strong moral compass), {\sc [setting]} develops the setting (e.g., most of the story takes places in the Hogwards School of Witchcraft and Wizardry, a fictional boarding school of magic for young wizards), and {\sc [plot]} outlines the plot elements (e.g.,~the boy  discovers he is the son of famous wizards and will attend Hogwarts School of Witchcraft and Wizardry). These planning agents target specific weaknesses observed in LLM-generated stories. Indeed, LLMs struggle writing compelling plots and consistent characters throughout long stories (see \Cref{sec:related_work}).

In addition to these content planning agents, we design five
\emph{writing agents}, each specialized in distinct elements of a
typical story structure: {\sc [exposition]}, {\sc [rising action]},
{\sc [climax]}, {\sc [falling action]}, and {\sc [resolution]}. We
adopt this structure since it is widely used and quite versatile
\citep{freytag1896freytag, pavis1998dictionary}, leaving other
narrative structures for future work. When generating in one go, LLMs
struggle to meet length requirements (e.g.,~specified in the prompt),
resulting in stories that are generally too short (see
\Cref{sec:results}).  Since our writing agents generate the final
output piecemeal, section by section, this results in longer outputs.

We model each of these agents as an LLM with a specific prompt template that formats the scratchpad into an appropriate prompt for each agent's subtask. Detailed scratchpad formatting and prompt templates for each agent are provided in \Cref{appendix:agent_prompts}. To coordinate between the different agents, we define a deterministic \emph{orchestrator} that first calls the planning agents as follows: {\sc [conflict]} $\to$ {\sc [character]} $\to$ {\sc [setting]} $\to$ {\sc [plot]}, then calls the writing agents: {\sc [exposition]} $\to$ {\sc [rising action]} $\to$ {\sc [climax]} $\to$ {\sc [falling action]} $\to$ {\sc [resolution]}. We choose to use a deterministic orchestrator for simplicity, given the strong narrative theory prior. %, that there is no human-in-the-loop, and that the agents are collaborative. 
In future work, more refined orchestrators with learned objectives can be explored, possibly expanding to a wider range of narrative structures. As a first step towards building adequate reward models for training such orchestrators, we investigate automated evaluation strategies for the long-form fiction writing task in  \Cref{sec:evaluation}.

\begin{figure}[t]
\begin{tcolorbox}[colback=white, colframe=my-green, coltitle=black, parskip=5mm,  title=
\textbf{Example Prompts}, breakable, halign=justify]
\vspace*{-.1cm}
  \begin{scriptsize}
{\setstretch{.1} 
Write a story about someone who is haunted by a ghost who wants to give business advice. This story should be around 2500 words. Don’t make it scary. The main character is trying to make her food truck popular, so she travels around the southwestern part of the country in her food truck to gain more popularity. After a long time on the road, she comes home to find a ghost. This ghost doesn’t want to scare her. He wants to give her business advice because he loved her food when he was alive. In the end, she accepts the help of the ghost.
}
   \tcbline
{\setstretch{.1} 
\begin{spacing}{.5}
Write a science fiction story about someone who is a time traveler and has dedicated everything in their life towards a goal, and now wonders if it was worth it. The story should be between 850 and 900 words. The story should begin with the main character waking up on a frozen tundra. He looks for shelter from the cold. He sees a dead wooly mammoth and realizes he traveled back to the ice age. The character should find shelter, and a predator is outside his shelter at night. The ending should not be happy.
  \end{spacing}
  }
%  \tcbline
%Write a story about a stranger coming to a small town and shaking up
%the order of things. The story should be about 950 words and be a
%science fiction story. The story should be framed with three old men
%gossiping about the stranger.  The story should be in the third person
%point-of-view. The stranger is found wandering in a rural town and is
%taken to a very small hospital. A doctor is called in to treat
%him. The stranger should recognize the doctor as an alien. The doctor
%tells the patient about the aliens' conspiracy to infiltrate
%earth. There should also be subtle hints that one of the old men is an
%alien. The ending should be scary.
\end{scriptsize}
\vspace{-.1cm}
\end{tcolorbox}
\vspace{-.2cm}
\centering
\caption{Prompts from the \tellmeastory\ dataset (corresponding stories are in \Cref{appendix:fictionstories}).\label{fig:prompts}}
\end{figure} 

\subsection{Synthetic Data Generation for Agent Training}
\label{sec:agent_training}

% \reinald{Create a figure to illustrate this backtranslation idea}

%\reinald{In charge of this section}

For each  agent, we explore zero-shot prompted and fine-tuned approaches, 
since a different degree of subtask specialization can be achieved through each approach. 
% -- the latter is useful given that each agent needs to specialize a certain subtask. 
Fine-tuning requires agent outputs, which are not readily available; planning agent outputs such as plot and setting are usually not provided in datasets, while writing agent outputs require the stories to be split into its constituent parts. Similar to previous work \citep{schick2022peer,narayan2023conditional,josifoski-etal-2023-exploiting}, we propose to generate synthetic outputs for these agents through distilled backtranslation.

Specifically, given as input writing prompts \emph{and} stories from a dataset (see Section~\ref{sec:dataset}), we zero-shot prompt a larger teacher LLM to (1)~generate the planning agent outputs, and (2)~segment the story into distinct parts (e.g.,~exposition, climax).
Note that unlike typical distillation methods, our task is more straightforward; all that is required is to reverse engineer the agent outputs from an existing story rather than generate them from scratch. 
The teacher LLM outputs are then used to generate synthetic training datasets for both planning and writing agents. 
% We use Gemini Ultra \citep{team2023gemini} as the teacher LLM. 
Detailed prompt templates are provided in
\Cref{appendix:synthetic_data_generation}.

%Given a labeled dataset of $(x, y)$ pairs of writing prompts and outputs, there is typically no labeled data for specialized agents' outputs. Following previous work , 
%we propose generating synthetic data to train the individual agents by distilling from an LLM. 

%For the planning agents' training data, we provide the LLM with the gold prompt and output from the training data. We then ask the LLM to backward generate the central conflict points, character descriptions, setting, and plot outline grounded on the provided gold data example. For the writing agents, we ask the LLM to segment the gold outputs into subsections corresponding to the exposition, rising action, climax, falling action, and resolution. Detailed prompt templates are provided in \Cref{appendix:synthetic_data_generation}. \reinald{Provide additional detail on the synthetic data generation process for the writing agents, and add the corresponding prompts to appendix} In our experiments we observed that this process yielded consistent and coherent training data for the different specialized agents (e.g., the generated character descriptions match the gold story). 

%\reinald{Comment on how this does not work as well without the gold data, the key here is backtranslation}

\subsection{\tellmeastory\ Dataset}
\label{sec:dataset}

% We formalize the fiction writing task as follows: given an initial writing prompt~$x$, generate output fiction story~$y$. 
Creative writing presents a particular challenge from a data collection perspective; it is not akin to any traditional annotation or evaluation task where a requester provides some input and some set of guidelines for marking up that input in a consistent manner.
%
%In this sense, most annotation and evaluation tasks entail the learning of some skills in terms of using a tool and analyzing inputs to transform inputs according to some standard that is defined to some extent on a per-task basis, excepting annotation tasks for which ground truth is defined by some externally defined standard, such as part-of-speech tagging. 
%
While standards exist for “good” writing, they evaluate the quality of writing across multiple interdependent and independent dimensions, all at once. In addition to this, the skill of writing really represents several skills that are learned over the course of a person's lifetime and educational experience.
%(Jenni to provide citation)
Furthermore, evaluating writing necessarily requires the subjective stance of the evaluator. %what one reviewer is pleased by in a piece of writing may not be enjoyable to another. 
%
%But perhaps most importantly, the traditional HIIT environment of each rater working more or less independently on a task one item at a time at a computer is hardly conducive to the creative process.
%

Taking into consideration all of these complexities, we collected  \tellmeastory\ through writing workshops to replicate the organic environment in which a collaborative writing process can take place. We provided a group of writers (28 in total) with broad instructions for quality based on collation of the Publication Manual of the American Psychological Association (currently in its 7th edition), the GRE/GMAT writing assessment rubrics, and various mass market style guides. 
%For the pilot workshop, we provided raters with writing prompts, which consisted of an identical set of root instructions with diverging sub instructions. % to aid in human evaluation. 
%For subsequent workshops, 
Writers created their own prompts, %to ensure topical relevance. %and %and divided raters by genre into short fiction, poetry, etc with leads 
%assigned to each workshop who had a relevant advanced degree or professional experience. For quality control, we followed the methodology of the workshop; 
wrote an initial draft, received feedback from peers, revised, and
then submitted to a workshop lead for a second round of feedback and
final approval. Workshop leads could ask for additional edits or
accept as is. Workshops lasted on average 3--4 weeks. %Raters had the
option of downtime between workshops or the opportunity to work in
another workshop if they desired to prevent burnout.  The average rate
of production for workshops generally reached no more than
\mbox{2--3}~writing samples per writer per week. We provide example prompts
in Figure~\ref{fig:prompts} and example stories in
\Cref{appendix:fictionstories}. The majority of the stories belong to
the genres of science fiction and fantasy but are also representative
of the following genres: horror, drama, comedy, adventure, and
folklore.
%We should also point out that stories often belonged to several genres (e.g., comedic horror or urban fantasy). We will add more discussion in the paper.

\begin{table}[t]
\begin{center}
\caption{\label{tab:comparison-datasets} Comparison of \tellmeastory\ against existing open-ended story generation benchmarks. We report statistics on the number of training, validation, and testing instances; Input/Target denote the average number of tokens in the input (aka prompt) and target text.}

\begin{tabular}{l|rrr|rr}
\toprule
\multicolumn{1}{c}{} &
\multicolumn{3}{c}{\emph{number of examples}} &
\multicolumn{2}{c}{\emph{avg. token length}}
\\
\multicolumn{1}{c}{Dataset} & \multicolumn{1}{c}{Training} & \multicolumn{1}{c}{Validation} & \multicolumn{1}{c}{Testing} &
\multicolumn{1}{c}{Input} &
\multicolumn{1}{c}{Target}
\\
\midrule
\textsc{WritingPrompts} \citep{fan-etal-2019-strategies}& 272,600 & 15,620 & 15,138 & 28 & 735\\
\textsc{RocStories} \citep{mostafazadeh-etal-2016-corpus} & 176,688 & 9,816 & 4,909 & 9 & 41\\
\textsc{ChangeMyView} \citep{hua-etal-2019-argument-generation} & 42,462 & 6,480 & 7,562 & 18 &  104\\
\textsc{WikiPlots}\tablefootnote{Available at: \url{https://github.com/markriedl/WikiPlots}} & 69,288 & 8,661 & 8,662 & 4& 195\\
\bf{\textsc{Tell Me a Story}} & 123 & 52 & 55 & 113 & 1,498 \\
% \textsc{WritingPrompts} \citep{fan-etal-2019-strategies}& 272,600 & 15,620 & 15,138 & 28.40 & 734.50\\
% \textsc{RocStories} \citep{mostafazadeh-etal-2016-corpus} & 176,688 & 9,816 & 4,909 & 9.02 & 40.72\\
% \textsc{ChangeMyView} \citep{hua-etal-2019-argument-generation} & 42,462 & 6,480 & 7,562 & 17.89 &  104.10\\
% \textsc{WikiPlots}\tablefootnote{Available at: \url{https://github.com/markriedl/WikiPlots}} & 69,288 & 8,661 & 8,662 & 3.88 & 194.72\\
% \bf{\textsc{Tell Me a Story}} & 123 & 52 & 55 & 112.93 & 1,498.17 \\
\bottomrule
\end{tabular}
\end{center}

\end{table} 

% Our dataset has three  splits, training / validation / test, with 123 / 52 / 55 examples, respectively. The inputs are on average 113 tokens long (42 standard deviation), while the targets have 1,498 tokens on average (540 standard deviation). 
Table~\ref{tab:comparison-datasets} compares \tellmeastory\ against commonly used story generation benchmarks. Our dataset is small in scale and thus not suited to training a model from scratch. Our prompts are more detailed compared to other benchmarks (see Input column) and the target stories are genuinely long (e.g,~double in size compared to \textsc{WritingPrompts}). Note that some of these datasets, although useful for system development, are not strictly speaking narratives. \textsc{WikiPlots} is a collection of  plots from Wikipedia rather  than stories, \textsc{RocStories} are five-sentence long common sense stories, and \textsc{ChangeMyView} contains pairs of posts and their counter-arguments.

%Jenni TODO: find out total number of raters, confirm description and numbers below

% \todo{Point out how this task is challenging, even for humans.}

% Dataset statistics: 
%           targets      inputs
% count   230.000000  230.000000
% mean   1498.173913  112.930435
% std     540.128931   42.321440
% min     548.000000   54.000000
% 25%    1058.000000   86.250000
% 50%    1428.000000  100.500000
% 75%    1924.750000  132.750000
% max    2868.000000  410.000000

\section{Experimental Setup}
\label{sec:experimental_setup}
% \fantine{In charge of this section}

%We describe details on the implementation of our framework and explore a number of 
%\Agentsroom\ variants. We also compare our system against baseline approaches. 

\textbf{Comparison Systems} 
The state-of-the-art approach for generating narratives consists of
generating the story in one go, either through zero-shot prompting
(see Figure~\ref{fig:prompts}) or fine-tuning, which we denote as
\baselinezs\ and \baselineft, respectively.  We also experimented with
more detailed instantiations of \baselinezs\ by instructing the model
to: (1)~generate the central conflict, characters, setting, and plot
before generating the story (\baselinezs\ plan); (2) reflect on the
central conflict, characters, setting, and plot according to detailed
guidelines, before generating the story (\baselinezs\ reflect);
self-reflection \citep{Madaan:ea:2023,Shinn:ea:2023} has been
previously explored to solve intricate tasks that could be challenging
for zero-shot prompting; both baselines (\baselinezs\ plan,
\baselinezs\ reflect) use the same detailed instructions provided to
our planning agents (see
Appendix~\ref{appendix:planning_agents_prompts}); (3)~generate a plan
automatically before generating the story in one call (\baselinezs\
decompose); in this case, plans are predicted without any
task-specific knowledge, story generation is decomposed into a series
of simpler sub-problems which are solved sequentially
\citep{yang-etal-2022-re3,Khot:ea:2023}; and (4)~generate a
plan for the story first followed by a second call in which the model
is instructed to generate the story based on the input prompt and the
plan (\textsc{2Stage} decompose).

\textbf{\Agentsroom\ Variants} We use the {\it plan+write} tag to denote the \Agentsroom\ variant with the writing and planning agents as previously described (see \Cref{sec:fiction_agents}). To  explore trade-offs between the different types of agents, we investigate two additional variants, {\it plan} and {\it write}, where we use only planning, or only writing agents, respectively. In the specific case of the {\it plan} variant with only  planning agents, we still need a writing agent to finalize the story, since planning alone does not result in a final story. Therefore, the {\it plan} variant includes a single simple writing agent,  which we denote as the {\sc [finalizer]}. The prompt template for the  {\sc [finalizer]} agent is provided in \Cref{appendix:agent_prompts}. We investigate both zero-shot and fine-tuned agents. For each \Agentsroom\ variant, we explore two settings, one with only zero-shot agents, and one with only fine-tuned agents,  denoted as \agentsroomzs\ and \agentsroomft, respectively. Since agents are called independently, it is possible to mix and match between zero-shot and fine-tuned agents, but we keep the two settings separate to derive clearer signal for each approach.

\textbf{Implementation}
For all comparison baselines and \Agentsroom\ agents, we use a Gemini
1.5 Flash\footnote{Available at: \url{https://cloud.google.com/apis}}
backbone, a lightweight and cost-efficient model that has demonstrated
good performance across a wide range of generative tasks
\citep{reid2024gemini}.  In particular, it features long context
capabilities (up to one-million tokens) which makes it suitable for
handling the scratchpad with multiple agents’ contributions. We use
input length out of \{1,024, 2,048, 4,096, 8,192\} tokens depending on
the length of the scratchpad and a target token length of~4,096. While
the outputs generated by the baseline systems are generally shorter
than what is requested in the original prompt (see
\Cref{sec:results}), we observe no improvements when increasing the
target token length. We hypothesize that the observed limits on output
lengths are likely due to the backbone model being trained on data
with mostly shorter outputs.

For the synthetic training data generation described in
\Cref{sec:agent_training}, we use Gemini Ultra\footnotemark[4]
\citep{team2023gemini} as the teacher model.
% Gemini 1.5 Pro\footnotemark[3] backbone.
Since our dataset contains only a small number of training examples,
we fine-tune our models (\baselineft\ and individual agents for
\agentsroomft) using LoRA \citep{hu2021lora}, a
computationally-efficient approach that updates only a small portion
of the model weights. We perform LoRA-tuning with rank 4 and a
learning rate of $1^{-6}$ (picked after a hyperparameter search
through \{$1^{-4}$, $1^{-5}$, $1^{-6}$, $1^{-7}$\}). We LoRA-tune for
250 steps with a batch size of 16, saving checkpoints every 20
steps. We then select the checkpoint with lowest loss on the
validation set.

\section{Evaluation}
\label{sec:evaluation}

We evaluate the quality of the generated outputs along several dimensions through human judgment elicitation and automated evaluation methods.

\subsection{Human evaluation}
\label{sec:human-eval}

%\mirella{In charge of this section}

We evaluate system output by soliciting pairwise preferences \citep{Louviere1990} along four dimensions, as well as an overall preference. We distill previous proposals  \citep{chakrabarty:ea:2024b,chhun-etal-2022-human} on how to  evaluate creative writing into the following criteria: 
\begin{itemize}

\item \textbf{Plot} --- Does the story have a recognizable structure, e.g., with a connected beginning, middle, and end? Does it exhibit  events and turns that move the plot forward without logical or conceptual inconsistencies?

\item \textbf{Creativity} --- Does the story have engaging characters, themes, and imagery?   Does it avoid overly cliched characters and storylines, unintentional tropes, and stereotypes? Does it include  original elements that were not explicitly mentioned in the prompt? 

\item \textbf{Development} --- Are the characters and settings contextualized with relevant details that allow the reader to understand their place in the story? Are appropriate levels of detail and complexity provided to lend the story a feeling of realness and believability? 

\item \textbf{Language Use} ---  Does the language used feel varied and rich? Does the story exhibit rhetorical, linguistic and literary devices  to create interesting effects? Does it avoid  bland or repetitive phrases? 

\end{itemize}

% \fantine{depending on space, comment on how these rubrics were refined to target specific weaknesses observed in LLM-generated stories. For instance, unlike previous work, we no longer include rubrics related to aspects such as grammatical correctness.}

The full instructions are reproduced in Appendix
\ref{appendix:instructions}. Participants are shown two stories and
asked to decide which one is better in each dimension. They can also
respond that the two stories are about the same. Participants are
allowed to rate up to five samples in one sitting, due to our task
being cognitively taxing and time-consuming. We assign samples to
participants following a Latin Square design, such that each
participant does not rate the same writing prompt more than once.  We
randomize the order in which the two stories are shown to mitigate
presentation order as a potential bias. We gather ratings for all
examples included in the \tellmeastory\ test set and compare outputs
from all \baseline\ and \Agentsroom\ variants (see
\Cref{fig:human_autorater_evals}); we also include the human-written
stories as an upper bound.  Our annotators were writers or had a
degree in related disciplines (e.g., literature).  We obtained a total
of 9,900 pairwise ratings which we converted into systems' relative
strengths using a Bradley-Terry model (\citealt{bradley1952rank}; see
Section~\ref{sec:automatic:eval}).  Inter-annotator agreement was
$\kappa = 0.46$ (\mbox{$p<0.01$}, $N=150$, $k=3$), as measured by
Fleiss' Kappa, which we interpret to be satisfactory given the
subjectivity of our task.

% \fantine{Do we need to include detail about the number of annotators and how much we paid them?}

\subsection{Automatic evaluation}
\label{sec:automatic:eval}

Many previous studies (see  \citealt{yang2024makesgoodstorymeasure} and the references therein) have highlighted the challenges
associated with evaluating narratives automatically. Metrics based on
lexical matching correlate poorly with human judgments
\citep{chhun-etal-2022-human,chakrabarty:ea:2024a} and do not
effectively measure story quality (e.g., is the story original and
well-written with plausible characters). In this work, we report
reference-based metrics, specifically Rouge-L \citep{lin-2004-rouge} and  BertScore \citep{bertscore:2020}, but also adopt
several surface-based metrics\footnotemark[1] intended to capture differences between
human writing and LLM-generated stories. Specifically, we compute
\emph{story length} to determine whether models are able to generate
long stories and  quantify  \emph{structural differences} between human and machine stories  (e.g.,~number
of sentences starting with an article or a pronoun). We also  measure the ratio of
\emph{unique words} in a story which gives an idea of creative
language use, and \emph{intra-} and \emph{inter-story trigram repetition}
\citep{yao2019plan,goldfarb-tarrant-etal-2020-content} which capture
diversity within a story and across stories (high inter-story repetition
suggests  models  generate similar  stories even when given
different prompts). Finally, trigram \emph{overlap with the prompt} is used to indicate 
whether models can creatively elaborate on the information provided.

In addition, we develop a LLM-based evaluator
\citep{liusie2023zero,liu2024aligning,zheng2024judging,bohnet2024long}
to perform \emph{side-by-side} comparisons of system output. We design
prompts targeting the same dimensions of story quality adopted in our
human evaluation.
%, namely development, originality, language use,
%plot, and overall (see Appendix for details). Using Goldfish we rated
%stories as what. 
%
%We found human judgments to correlate significantly
%with Goldfish ratings, yielding an overall Pearson's $\rho$ correlation
%of 0.41~across dimensions (\mbox{$p<0.01$}). 
%
%In addition to human elicitation, we use a number of automated evaluation methods, investigating LLMs as autoraters and computing surface-form metrics. 
%
%\paragraph{LLM Autorater}
%
%\reinald{Write about Goldfish metrics}
%
%\cite{bohnet2024long} reported that LLMs, specifically Gemini 1.5 Pro \citep{} and GPT-4 Turbo \citep{}, can be effectively used as an autorater. That is, provided they have sufficient context, LLMs can be good \textit{side-by-side} autoraters of text, with good self-agreement (i.e., they are consistent) and human agreement (i.e., they are effective). In this paper, we follow \cite{bohnet2024long} with two motivations. Firstly, we want to test whether their hypothesis holds in the Fiction Writing Task, which is more complex than the QA task presented in their paper. And secondly, we want to develop a fully-automated autorater that we can use to compare and rank different systems.
%
Specifically, we adapt the  evaluation criteria described in Section~\ref{sec:human-eval} into a prompt template shown in Appendix~\ref{appendix:sxs_template}.
This template asks the evaluator for a detailed assessment of the two stories presented, followed by a final conclusion, which is then parsed to obtain preference scores for each dimension.
% of story quality: 
% plot, creativity, development, language use, and overall. 
We provide an example usage in Appendix~\ref{appendix:sxs_template}.
% Given $N$~system outputs for each of $M$~input prompts, we obtain all possible pairs of outputs for each input, and apply them to the template (the order of outputs is shuffled), producing $M \times N \times (N-1)/2$ different prompts for the LLM.
Given $N$~system outputs for each of $M$~input prompts, we evaluate all possible (unordered) pairs of outputs for each input (while shuffling the order in which the outputs are presented), producing $M \times N \times (N-1)/2$ different pairwise ratings.
Finally, we obtain a wins matrix $W$ where $w_{i,j}$ is the number of times system $i$ wins over system $j$. This matrix is then used to obtain the systems' relative strengths after fitting a Bradley-Terry model \citep{bradley1952rank}.
We use Gemini 1.5 Pro\footnotemark[4] as our LLM evaluator, as suggested in \cite{bohnet2024long}.

%\reinald{Fill in appendix sxs template}
%We report the self-agreement and human agreement results in Section~\ref{sec:results}.

\section{Results}
\label{sec:results}

\begin{table}[t]
\begin{center}
\caption{Comparison between human and model generated stories using
  automatic metrics (\tellmeastory\ test set): \#words (average number
  of words per story), \#para (average number of paragraphs per
  story), Article (proportion of sentences starting with an article),
  Pro (proportion of sentences starting with a pronoun), Unique
  (percentage of unique words), Intra (intra-story trigram
  repetition), Inter (inter-story trigram repetition), Overlap
  (proportion of trigrams overlapping with the prompt). We also report
  two reference-based metrics, Rouge-L and BertScore. \textsc{AR}
  abbreviates \textsc{Agents' Room} systems; subscripts $_{ZS}$ and
  $_{FT}$ respectively refer to zero-shot and fine-tuned.}
\label{tab:surface-metrics} 

\begin{tabular}{@{}l@{~~}r@{~~}c@{~}c@{~}c@{~}c@{~}r@{~~}c@{~}c@{~}c@{~}c@{}}
\\
\toprule
Models   & \#words & \#para & Article & Pro &  Unique & Intra & Inter & Overlap &
Rouge & BertSc \\ \midrule
  \baselinezs & 1,207 & 32.24 & 12.74 & 40.45 &  44.57 & 28.78 & 33.35
  & .0034 & 20.71 & .8152 \\
  \baselinezs~plan&  1,130 &27.24 &15.34 & 42.25 &45.93 &	23.59
  &	29.41& 	.0027 &	20.58& 	.8173\\
  \baselinezs~reflect& 1,126 &	28.62 &	13.79 &	40.96& 	45.85& 	23.68
  &	23.95& 	.0032 &	20.36 & 	.8152\\
  \baselinezs~decompose & 965 &21.25&21.36& 39.62 &	45.49 &	31.98
  &44.10& 	.0034 & 	19.41 &	.8067\\
  \baselineft & 1,193 & 32.25 & 12.58 &   43.39 & 44.02 &28.21 & 31.31
  & .0036 & 20.73 & .8138 \\ 
\textsc{2Stage} decompose &  	1,090 &	24.82 &	15.59 &	42.15& 	44.50&
21.54 &	24.26 &	.0031 	&20.35 &	.816 \\\hline
  \agentsroomzs~plan & 926 & 20.95 & 13.82 & 40.68 & 43.88 & 29.70 & 33.49 & .0017 & 19.58 & .8119 \\
  \agentsroomzs~write & 3,278 & 63.80 & 25.32 & 39.91 &34.97 & 47.50 & 44.09 & .0022 & 17.34 & .8103 \\
  \agentsroomzs~plan + write  &  3,034& 58.65 & 15.97 & 41.43 &  35.05 & 44.73 &
  43.25 & .0022 & 17.57 & .8123 \\
  \agentsroomft~plan & 856 & 21.05 & 18.02 & 39.29  & 44.65 & 23.85 & 28.05 & .0027 & 19.24 & .8146 \\
  \agentsroomft~write & 3,129 & 61.90 & 17.45 & 44.80 & 36.35 & 46.39 & 42.39 & .0021 & 17.53 & .8150 \\
   \agentsroomft~plan + write & 3,006 & 56.85 & 17.52 & 43.03 & 34.30 & 46.31 &
   41.60 & .0019 & 17.60 & .8152 \\ \hline
     Humans & 1,439 & 32.91 & 10.01 & 32.37 & 50.35 & 15.53 & 19.24 &
  .0020 & --- & --- \\\bottomrule
\end{tabular}
\end{center}
\end{table}

Table~\ref{tab:surface-metrics} compares human and model generated
stories using surface- and reference-based metrics. As far as story
length is concerned, we observe that \baseline\ stories are slightly
shorter than human ones, while planning models are shortest
overall. However, models which include writing agents produce
considerably longer stories (by a factor of two) with more dialogue as
suggested by the increased number of paragraphs.  We also find machine
stories to be more generic in their sentence structure as evidenced by
the higher proportion of stories which start with an article or
pronoun.  Human-written stories are also more diverse (less
repetitive) as shown by the higher ratio of unique words and less
repeated trigrams (Inter and Intra in
Table~\ref{tab:surface-metrics}).  The most repetitive models are also
the ones that produce the longest stories. In terms of overlap with
the prompt, we find \Agentsroom\ systems to copy least, at a rate
similar to that of human writers. Rouge-L rewards the \textsc{e2e}
systems most, as they least deviate from the prompt gold standard
stories, while BertScore is not very discriminating, equally
preferring the simplest (\baselinezs) \emph{and} most complicated
system (\agentsroomft~{\it plan+write}). Examples of stories written by humans and machines can be found in Appendix~\ref{appendix:fictionstories}.

\begin{figure}[t]
\centering
% \begin{tabular}{@{}l@{\hspace{-3cm}}l@{}}
% \hspace{.6cm}a. Human-based ranking & \hspace{.6cm}b. LLM-based ranking\\
% \includegraphics[width=0.7\textwidth]{human_eval}
% & 
% \includegraphics[width=0.7\textwidth]{autorater_eval}
% \end{tabular}

% [trim={left bottom right top},clip]
\includegraphics[width=\linewidth, trim={7 7 7 0}, clip]{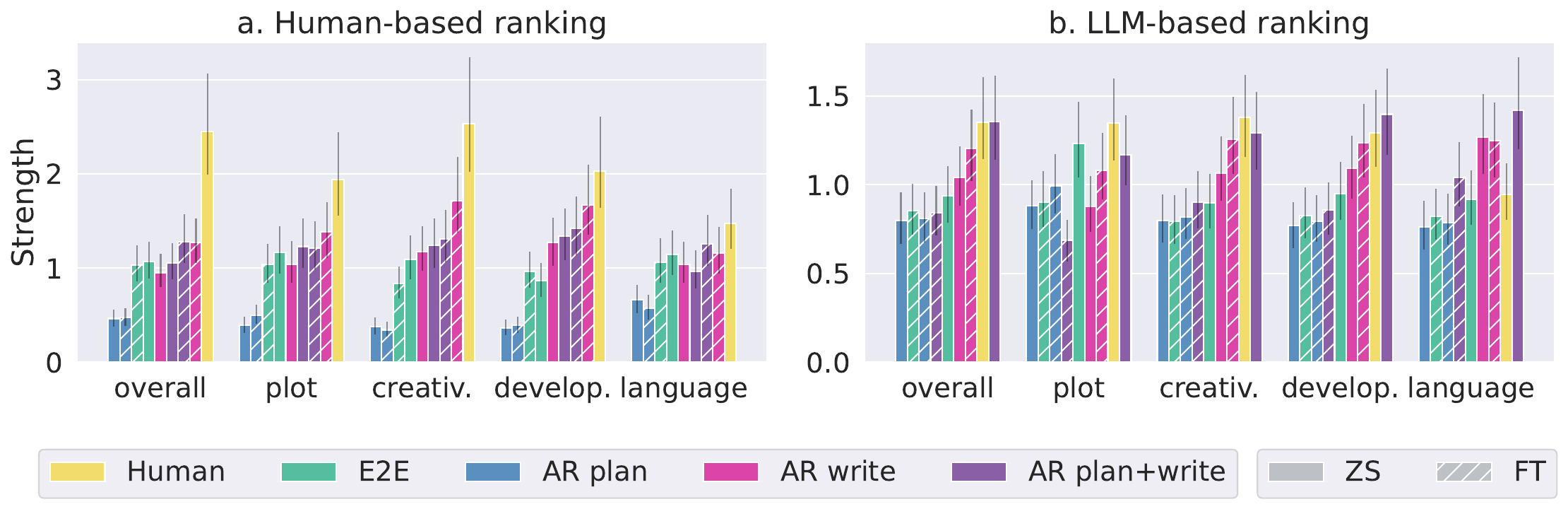}
\vspace{-.4cm}
\caption{\label{fig:human_autorater_evals} Overall system ranking across dimensions of plot, creativity, development, and language, according to human ratings~(a) and a LLM-based evaluator~(b).}
\end{figure}

\Cref{fig:human_autorater_evals} reports system rankings obtained from
human judgments and the LLM evaluator. For the sake of brevity,
we omit  baselines shown to underperform in
Table~\ref{tab:surface-metrics} (i.e.,~\baselinezs\ with \emph{plan,
deflect, decompose} and \textsc{2Stage} \emph{decompose}) but include
these results in Appendix~\ref{sec:additional-results}.

\textbf{Human-written stories are preferred overall} As shown in~\Cref{fig:human_autorater_evals}a, human judgments reveal a
performance gap between machine writers (\baseline\ and \Agentsroom)
and professional writers, a finding that is in line with
\cite{chakrabarty:ea:2024a}. We observe this gap across all
dimensions, but we note that it is smaller in the language use
dimension. This
result suggests that while machine-generated stories still fall short
in terms of compelling plots and unique ideas, LLMs, in their current
state, may be useful as writing assistants. To ensure that the preference towards human stories is not
merely due to them being longer, we computed the proportion of
pairwise comparisons for which our human raters preferred the longer
story overall (excluding ties) and found it to be around~0.51. 
% As such, future work could have humans and LLMs write together.

\textbf{\Agentsroom\ outperforms baseline systems} Across all
dimensions, our participants prefer \Agentsroom\ stories with writing
agents over those produced by baseline systems, with the
\agentsroomft~{\it write} and \agentsroomft~{\it plan+write} systems
performing best. Aside from rating the stories, participants had the option to leave feedback on their quality; we provide samples of this feedback  in Appendix~\ref{sec:additional-results}. {\sc AR}~{\it plan} variants do not perform that
well, most likely due to the single {\sc [finalizer]} agent being too
simplistic to make good use of the planned elements provided in the
scratchpad. We note that fine-tuned agents yield better results over
zero-shot ones, which shows that generating synthetic data by
back-translating from gold standard outputs (see
\Cref{sec:agent_training}) is an effective strategy for training
specialized agents for different subtasks. Finally, we observe similar
trends with smaller scale models (see
Appendix~\ref{sec:additional-results} for additional results).

\textbf{The LLM evaluator agrees with humans and itself} The LLM-based rankings in~\Cref{fig:human_autorater_evals}b reveal similar tendencies to human ratings. The LLM overall prefers human stories and those generated by the {\sc AR}~{\it plan + write} system against all other model variants, even though it does not discriminate very strongly between those two. LLM-based judgments of story quality  correlate significantly
with human ratings across all dimensions, both  by systems (Spearman's rank correlation $\rho=0.62$; \mbox{$p<0.01$}, $N=45$) and  by items ($\rho=0.41$; \mbox{$p<0.01$}, $N=9,900$). The  LLM and humans have the highest agreement when assessing story development ($\rho=0.83$, \mbox{$p<0.01$}) and creativity ($\rho=0.85$, \mbox{$p<0.01$}). Similarly to the findings in \cite{bohnet2024long}, we also find that the LLM evaluator scores are consistent: 90.2\% of the time the LLM prefers the same story in a second run, when the stories are presented in the opposite order. 

%This shows the potential of our evaluation scheme for longer-form outputs. Point out the benefits of the fine-grained evaluation system.

%\reinald{Adding some of my self-agreement experiment results here.}

%To check whether or not the LLM autorater is consistent, we conducted several self-agreement tests. Firstly, we shifted around the order of the systems and reran the autorater to see if the order changes the results

% \section{Discussion} 
% Future work can have human and machine-agents collaborate. With improved evaluation, it would become possible to train more complex orchestrators.

\section{Conclusion} 
We propose \Agentsroom, a general framework for multi-agent collaborative writing, and describe its instantiation for the long-form fiction writing task. Drawing inspiration from narrative theory, we  decompose the complex writing task into subtasks tackled by specialized agents. To illustrate this framework, we present \tellmeastory, a high-quality dataset of prompts and long-form stories collected through multiple rounds of writing workshops with human participants. We show that \Agentsroom\ generates stories that are preferred by human evaluators over those produced by baseline systems. Moreoever, we demonstrate effective training stategies for developing specialized agents by leveraging  synthetically-generated data. We introduce a human evaluation framework for evaluating long-form narratives across several dimensions, and an LLM-based evaluator that correlates significantly with human raters. With improved automated evaluation, future work can explore more sophisticated orchestrators, including the development of reward models and learning objectives for such orchestrators.

% Future work can have human and machine-agents collaborate. With improved evaluation, it would become possible to train more complex orchestrators.
 
%\begin{tikzpicture}
%\tile[green!20]{1}{1}{0}{0}%
%\tile[red!20]{1}{-1}{0}{-1}%
%\tile[red!20]{1}{-1}{0}{-1}
%\tile[red!20]{1}{-1}{0}{-1}%
%\tile[red!20]{1}{-1}{0}{-1}%
%\tile[yellow!20]{-1}{1}{0}{-1}%
%\tile[yellow!20]{1}{-1}{0}{-1}%
%\tile[yellow!20]{-1}{1}{0}{-1}%
%\tile[yellow!20]{1}{0}{0}{-1}
%\end{tikzpicture}

%\raisebox{1cm}[0pt]{
%begin{tabular}{lllllllll}
%& \hspace{.7cm}3 & \hspace{.4cm}2 & %\hspace{.4cm}1 & \hspace{.4cm}4 & %\hspace{.3cm}d & \hspace{.4cm}a & %\hspace{.4cm}b & \hspace{.5cm}c 
%\end{tabular}
%}

\subsubsection*{Ethics Statement}
There are a number of ethical considerations when using generative language models. While the work we present here makes a step towards improving the quality of text generation systems, it is important to note that current systems are still far from perfect in this respect and may make mistakes. In particular, generative models may perpetuate biases present in their training data. Even when writing fiction, the models may inadvertently amplify societal biases or reinforce stereotypes, leading to the production of biased content. Therefore, the generated outputs should be meticulously verified and used with caution.

\subsubsection*{Reproducibility Statement}
For reproducibility, we release the \tellmeastory\ dataset on which we conduct our experiments, complete with its train, validation, and test splits, as described in \Cref{sec:dataset}. We specify the model backbones, implementation details, and where to access the checkpoints in \Cref{sec:experimental_setup}. All prompt templates and scratchpad formatting templates are provided in Appendix. For the evaluation, we provide the exact rater instructions in \Cref{appendix:instructions} and the LLM evaluator prompts in \Cref{appendix:sxs_template}.

\bibliography{custom}
\bibliographystyle{iclr2025_conference}
\newpage
\appendix
\section{Examples from \tellmeastory\ Dataset and System Output}
\label{appendix:fictionstories}

% green!30!gray
\begin{tcolorbox}[colback=white, colframe=my-green, coltitle=black, parskip=5mm, title=
\textbf{Example Prompt}, breakable]
\scriptsize
  Write a story about someone who is haunted by a ghost who wants to give business advice. This story should be around 2500 words. Don’t make it scary. The main character is trying to make her food truck popular, so she travels around the southwestern part of the country in her food truck to gain more popularity. After a long time on the road, she comes home to find a ghost. This ghost doesn’t want to scare her. He wants to give her business advice because he loved her food when he was alive. In the end, she accepts the help of the ghost.
\end{tcolorbox}

\begin{tcolorbox}[colback=white, colframe=my-green, coltitle=black, parskip=5mm, title=
    \textbf{Human-written Story}, breakable]
  \scriptsize
  \begin{spacing}{.8}
“Finally, I can sleep in my own bed for once,” Rita said. \\\par
        She plopped down on her bed in an attempt to soak in its comfiness. But then there was a knock at the door, in her mind she said “Why!” \\\par
        Rita dragged herself from her bed to check to see who it was, but when she got to the door, there was no one there. She looked around to see if anyone was around, but there was nothing. She wondered if anyone with kids had moved in while she was away. \\\par
        Disregarding the thought she went back into her apartment and went back to her bed. She didn’t care to remove her shoes or her coat, that was just how tired she was. Rita had been on the road for the past five months trying to drum up enough money and support to start her own restaurant. Currently, she had been operating out of a food truck that she had started a year ago. She went from city to city every week, trying to find the perfect place that garnered enough foot traffic and business. \\\par
        After five months of going around the southwestern part of the United States, driving, cooking, serving, and advertising her business wherever she went, she decided that it was time to go back to home base and rest up a bit before she went back to the grind. It was three o’clock in the morning when she finally got up to get a little more comfy. \\\par
        She went into the kitchen where she had left her bags since it was the first place you reached once you entered the two-bedroom apartment. Rita didn’t have the energy to drag them along with herself to her room at the back of the apartment. \\\par
        Once she got to the kitchen she noticed that the flyers she had made for her food truck were scattered on the ground. She picked them up, but what she found on the flyers when she looked at them scared her. \\\par
        Rita could have sworn she saw something moving in the flyer; she admitted the flyers had an intricate design that looked like an optical illusion but what she saw was not a part of the design. Taking a second look, she did not see what she thought she saw.  \\\par
        “Nope nope nope, didn’t see a thing.” Rita decided that she just needed more rest after all it was three in the morning.
        Back in her room, Rita settled to get back in bed when she heard the knocking, but this time it was coming from her bathroom. She was definitely freaking out now. She grabbed the closest thing she could use as a weapon. A wooden crate, with a faded beer label on it, was all there was in her line of sight. \\\par
        She tiptoed her way to the bathroom, with the crate held above her head ready to strike anything that moves. She pushed the door open with her foot, and saw a man standing in her bathroom. She swung the crate, but it passed through the man, and she fell backwards. Seeing this, she wanted to close the door, but she couldn’t since she had already fallen to the floor and was backing up from the entrance to the bathroom. 
        Gathering the courage to approach the bathroom again, Rita saw nothing but a note on her cheval mirror that read, “Please don’t be afraid of me, I just want to help you.” The bathroom was ice-cold all of a sudden. Still standing in the doorway, Rita saw the man again, and this time he waved. After the initial shock of seeing him the first time she just waved back. \\\par
        Taking a deep breath in and exhaling, Rita said, “Okay, how is it that you want to help me?” She thought she was going crazy or something because she intently waited for this transparent being to communicate with her. She refused to admit it was a ghost, so she just waited. \\\par
        Finally, the being began writing on the mirror again. It was freezing cold in the bathroom so the mirror appeared to have frost on it. He explained in writing, “this would go faster if you would allow me near you.” \\\par
        Reluctantly, she agreed, the man approached her, and she decided to close her eyes as if she could somehow pretend this wasn’t happening. He touched her shoulder, all of a sudden Rita was hearing him in her mind. \\\par
        “Hi, I’m Jeff. I must say your cooking was the absolute best when I was alive.” Shocked by his admission, Rita opened her eyes to look at him. \\\par
        Jeff stood there with a smile on his face and continued, “I had the pleasure of trying your southern fried cabbage the day I died. It was truly the most fitting last meal.” 
        Amazed by his candor, Rita tried to remember when she may have seen this guy at one of the many locations she had been to in the last five months, but she couldn’t place him. Giving her time to digest the information, Jeff continued, “You won’t remember me because I wasn’t the one that picked up the order, it was my wife. But anyway, let’s get back to the point here. I want to help you with your food truck venture.”
        Rita took a minute and thought, “How would a dead man help me with this?” Jeff smirked at her because he knew she was thinking this was crazy. She shrugged her shoulders and gave in, asking, “Okay, how do you suppose you are going to help, because I don’t think having a ghost doing tricks will attract people.” \\\par
        Jeff shook his head and just looked at Rita until she finished her muttering. \\\par
        “First and foremost, I don’t do tricks. Second, you need an upgrade girlie girl. Who uses printed out flyers anymore in this technological world? Also, that little design you got on those flyers — scrap it. I got lost for hours in that thing, trying to get your attention.” \\\par
        Before Jeff could finish, Rita muttered under her breath, “I knew I saw something on those flyers.”\\\par
        Jeff continued to speak, “Rita, you need to expand even more than what you have. I’m not saying to expand your territory. I’m saying to reach more people you need to get on those food delivery apps, or at least create a website that will allow people to place orders.” Rita listened on, and she knew what Jeff was saying was true. She was stuck in a time long past when technology was not as widespread. \\\par
        After Jeff finished what he had to say, Rita asked, “So, if I do all of this, will you let me sleep in peace and stop knocking on everything to get my attention? If you have something to say just leave a note on the mirror.” \\\par
        Laughing at Rita’s acceptance of him and the things he set out for her to do. He gave her some people to contact about setting up some things as far as the delivery service went. Then he allowed her to go to sleep. \\\par
        “See you later Jeff, we will continue this when I wake up.” Rita said as she closed the door to her room and finally got into her bed and fell asleep. 
        \end{spacing}
\end{tcolorbox}
\begin{tcolorbox}[colback=white, colframe=gray!30!white, coltitle=black, parskip=5mm, title=
    \textbf{System output: \baselinezs}, breakable]
    \scriptsize
    \begin{spacing}{.4}
    The desert sun beat down on Amelia’s food truck, “The Wandering Wok,” as she flipped a sizzling stack of tofu, her brow furrowed in concentration. For the past year, she’d been crisscrossing the Southwest, her truck a mobile testament to her culinary dreams. She’d started with a simple concept: innovative vegetarian takes on classic Asian dishes, fueled by fresh, local ingredients. It was a hit in her hometown, but she craved more. Now, she was pushing the limits, trying new recipes, adjusting her menu, even experimenting with different locations.  \\\par

Still, the crowds weren't as big as she hoped. Sometimes, there were just a few regulars, their faces familiar and comforting, but not enough to sustain her. The loneliness of the road was starting to get to her. The constant hustle, the uncertainty, the lack of a real home – it was wearing her down.  \\\par

One evening, after a particularly disappointing day in Sedona, Amelia packed up her truck, feeling a familiar pang of disappointment. The drive back to her hometown in Arizona was long and quiet. By the time she reached her childhood home, a cozy bungalow with a sprawling backyard, she was exhausted.  \\\par

Stepping inside, Amelia was greeted by the comforting scent of dust and memories. Her parents had passed years ago, leaving the house to her. She’d kept it as a haven, a place to return to when the road got too much. As she unpacked her bags, a strange sensation washed over her. It wasn’t fear, but a sense of…presence.  \\\par

A shiver ran down her spine as she turned, her gaze falling on a figure standing in the kitchen doorway. It was a man, tall and wiry, with a mischievous glint in his eyes. He was transparent, a faint shimmer of light outlining his form. A ghost.  \\\par

Amelia’s initial fear quickly dissipated, replaced by a strange sense of calm. The ghost, dressed in a faded, old-fashioned suit, smiled warmly. “Don’t be alarmed,” he said, his voice a gentle whisper. “I’m not here to scare you.”  \\\par

His words, though spoken in a spectral voice, held a reassuring warmth. Amelia, still reeling from the shock, simply stared at him. \\\par

“I’m Jacob,” the ghost continued, his smile widening. “And I’ve been watching you.” \\\par

Amelia’s eyebrows shot up. “Watching me?” \\\par

“Yes,” Jacob chuckled. “Your truck, ‘The Wandering Wok.’ I loved your food. I used to come by every week when you were still in town.” \\\par

He gestured toward the kitchen table, where a plate sat, adorned with a single, perfectly-formed veggie spring roll. It was a dish Amelia had perfected years ago. \\\par

“I’ve been following you all over the Southwest, you know,” Jacob said, his voice tinged with a hint of pride. “I’ve seen your struggles, your successes, your…lack of customers.” \\\par

Amelia, still processing the situation, felt a blush creep up her cheeks. “Well, I’m trying,” she mumbled, her voice barely above a whisper. \\\par

“You’re trying too hard,” Jacob said, shaking his head gently. “You’re trying to be everything to everyone. You need to focus, find your niche.” \\\par

His words hit Amelia like a jolt of electricity. It was true. She’d been trying to please everyone, offering everything from vegan pho to spicy Korean bibimbap, hoping to attract a wider audience. But in the process, she’d lost her own unique voice. \\\par

“You have a gift, Amelia,” Jacob continued, his gaze intense. “You make food that’s both delicious and beautiful. But you’re hiding it behind a facade of…well, too much.” \\\par

Amelia felt a wave of frustration wash over her. “It’s not that easy,” she said, her voice tight. “Competition is fierce. Everyone's trying to be the next big thing.” \\\par

Jacob chuckled, a soft, ethereal sound. “Competition is good, Amelia. It pushes you to be better. But you can't compete if you don't know what you stand for.” \\\par

He paused, his eyes twinkling. “You need to find your story, Amelia. What makes you different? What makes your food special?” \\\par

Amelia thought about it. Jacob was right. She’d been so focused on the bigger picture, on the grand vision of a successful food truck empire, that she’d forgotten the simple joy of creating.  \\\par

“My grandma,” she said, the words coming out in a rush. “She taught me how to cook. Her recipes, her stories, her love for food…that’s what makes my food special.” \\\par

Jacob smiled. “See? You have a story. A beautiful, personal story that people will connect with. You just need to tell it.” \\\par

He went on, sharing his own story, a tale of a young man who had come to Arizona from a small town in the Midwest, seeking his fortune. He’d worked hard, saving up every penny to open a small diner, a place where he could share his love for food with the community. He’d been happy, he said, until a sudden illness had taken him away. \\\par

“I never got to share my dream with the world,” Jacob confessed, his voice tinged with sadness. “But you, Amelia, you can. You have the talent, the passion, the story. All you need is to find your own path.” \\\par

He looked at her, his expression serious. “You need to create a menu that reflects your heritage, your grandma’s recipes, your own unique twist. You need to tell your story, Amelia. And you need to do it with passion, with love, with the same dedication your grandma had.” \\\par

Amelia felt a surge of inspiration. Jacob’s words, though spoken by a ghost, resonated deeply within her. She’d been so lost in the pursuit of success that she’d forgotten the essence of what made her food special.  \\\par

“Thank you,” Amelia whispered, her voice choked with emotion. “I…I think I know what I need to do.” \\\par

Jacob smiled, a gentle glow radiating from him. “I knew you would,” he said. “Now go, Amelia. Go out there and make your grandma proud.” \\\par

As quickly as he had appeared, Jacob vanished, leaving Amelia alone in the quiet kitchen. The plate with the spring roll was still on the table, a testament to his presence, his love for her food.  \\\par

Amelia spent the next few weeks reworking her menu. She incorporated her grandma’s recipes, her own personal touches, her love for fresh, local ingredients. She created a menu that told a story, a story of family, tradition, and passion.  \\\par

She also decided to rebrand her truck. “The Wandering Wok” was gone, replaced by “Grandma’s Kitchen,” a name that resonated with warmth and authenticity. \\\par

The first time she took her revamped truck out, Amelia felt a nervous excitement mixed with a sense of peace. She had found her voice, her story. As she set up her truck in a bustling market in Phoenix, she felt a familiar presence, a gentle warmth surrounding her.  \\\par

Jacob was there, a silent observer, his form shimmering faintly in the afternoon sun. He smiled at her, his eyes filled with pride. \\\par

“This is your story, Amelia,” he whispered. “Tell it with your heart, and the world will listen.” \\\par

Amelia took a deep breath, her heart filled with gratitude. She had found her path, her purpose, thanks to a ghost who simply wanted to share his love for food and his belief in her. And as she started cooking, her hands moving with newfound confidence, she knew that Jacob would be with her, cheering her on, every step of the way.  \\\par

The crowds came, drawn in by the aroma of her grandma’s recipes, the warmth of her story, the passion that radiated from her every move. They tasted the love, the care, the heritage woven into each dish.  \\\par

Amelia’s journey had taken a detour, a detour that led her back to her roots, to her grandma’s kitchen, to the true essence of her culinary dreams. And as she saw the smiles on the faces of her customers, she knew that Jacob, the ghost with a passion for food, was finally at peace. His story, his dream, was living on, through her. 
\end{spacing}
\end{tcolorbox}    
    
    \begin{tcolorbox}[colback=white, colframe=gray!30!white, coltitle=black, parskip=5mm, title=
    \textbf{System output: \Agentsroomzs}, breakable]
    \scriptsize
    \begin{spacing}{.9}
    The desert sun beat down on the cherry-red food truck, baking the metal siding until it shimmered like a mirage.  Lily, her forehead creased with a mixture of sweat and frustration, stared out at the empty parking lot, her reflection wavering in the heat haze.  Another day, another disappointing turnout.  The farmers' market had been slow, and the lunchtime crowd had been nonexistent.  The southwest was vast and beautiful, but the road to success seemed endlessly long.\\\par
    
Lily sighed, running a hand through her messy brown hair.  She'd dreamt of owning a food truck since she was a kid, watching her grandmother whip up culinary magic in their tiny kitchen.  The smells of her grandmother’s cooking, the warmth of the family gathered around the table, the joy in her grandmother’s eyes as she presented each dish...it was a memory Lily cherished.  She wanted to share that same joy, that same passion, with the world.  But the reality was far from the dream.  The food truck, her pride and joy, was becoming a burden.  The money was dwindling, and the pressure was mounting.\\\par

She’d spent the last six months crisscrossing the southwest, driving from one bustling market to another, her truck a beacon of hope in a sea of uncertainty.  She’d tried everything – catchy slogans, bright signage, even offering free samples.  But nothing seemed to work.  The competition was fierce, and the customers were fickle.  Lily felt like she was losing her grip, her confidence slowly slipping away. \\\par

Exhausted and discouraged, she decided to head home.  Her childhood home, nestled in a dusty corner of the small town of Tombstone, Arizona, was a place of comfort and memories.  Maybe a change of scenery, a little time away from the relentless grind of the road, would help her clear her head.  She needed a break, a chance to regroup and re-evaluate her strategy.\\\par

The drive back was quiet, the only sound the hum of the engine and the occasional sigh of the wind whipping through the open windows.  As she passed through the familiar landscape, the desert wildflowers swaying in the breeze, the towering red rock formations casting long shadows across the parched earth, a sense of melancholy settled over her.  The southwest, once a symbol of adventure and opportunity, now felt like a reminder of her own failures.\\\par

The sun had begun its descent, painting the sky with shades of orange and purple, as Lily finally turned onto the familiar street leading to her childhood home.  The adobe house, with its terracotta tile roof and sun-baked walls, stood bathed in the golden light, a testament to time and resilience.  It was a place of warmth and comfort, a place where she felt safe and loved.  But as she pulled into the driveway, a shiver ran down her spine.  A sense of unease, a feeling of being watched, crept into her heart.  It was as if the house itself was holding its breath, waiting for something.\\\par

Lily shook her head, trying to dismiss the feeling.  It was just her imagination, she told herself.  She was tired, stressed, and probably just a little bit paranoid.  But as she stepped out of the truck, the feeling intensified.  The air around her seemed to crackle with an unseen energy, and a faint, cold breeze sent a shiver down her spine.  She glanced around, but saw nothing.  The street was empty, the only sound the chirping of crickets and the distant howling of coyotes.\\\par

She quickly shrugged off the unsettling feeling, grabbing her bags and heading towards the house.  The front door, a heavy wooden structure adorned with a rusty brass knocker, creaked open as she pushed it, revealing a dark and dusty interior.  The scent of old wood and forgotten memories hung in the air, thick and heavy.\\\par

As she stepped inside, a sense of unease settled over her.  It wasn't just the dust and the shadows that made her uneasy, but an almost tangible feeling of presence, a sense of being watched.  She switched on the light, illuminating the worn-out furniture and the faded photographs on the walls.  The house, once a haven of childhood joy, now felt strangely empty, as if its heart had been ripped out.\\\par

Lily took a deep breath, trying to calm her nerves.  She was home, and that was all that mattered.  She would unpack, make a cup of tea, and then try to figure out her next move.  She needed a plan, a way to get her food truck back on track.  But as she started to unpack, a faint shimmer caught her eye.  A flicker of movement in the corner of the room, barely visible, like a heat wave shimmering in the summer sun.  Lily froze, her heart pounding in her chest.  She blinked, rubbing her eyes, but the flicker remained, growing stronger, taking shape.\\\par

It was a man, a tall, slender figure with a kind face and a gentle smile.  But he wasn't solid.  He was translucent, like a wisp of smoke, and his clothes were faded and worn, as if he belonged to a different time.  He stood there, watching her, his eyes filled with a mixture of curiosity and concern.\\\par

Lily's breath caught in her throat.  Her mind raced, trying to make sense of what she was seeing.  This couldn't be real, she told herself.  She was exhausted, stressed, and probably hallucinating.  But the figure didn't disappear.  He simply stood there, watching her, his presence as real as the dust motes dancing in the sunlight streaming through the window.\\\par

And then, he spoke, his voice a gentle baritone, as comforting as a warm fire on a cold night.  "Welcome home, Lily."\\\par

Lily's voice caught in her throat.  She couldn't speak, couldn't even breathe.  The man, the ghost, stood there, his translucent form shimmering in the afternoon light.  He seemed to be smiling, his eyes crinkling at the corners as if in amusement, but the sight of him sent chills down her spine.\\\par

"Don't be afraid," the ghost said, his voice a gentle murmur.  "I mean you no harm."\\\par

Lily managed to swallow, her throat dry as desert sand.  "Who...who are you?" she stammered, her voice barely a whisper.\\\par

"My name is Mr. Henderson," the ghost replied, his voice a comforting baritone.  "I used to live here, in this very house."\\\par

"You...used to live here?" Lily repeated, her mind struggling to grasp the impossible.  "But...you're...you're a ghost."\\\par

Mr. Henderson chuckled softly, a sound like wind chimes tinkling in the breeze.  "Yes, I am.  But don't be afraid.  I'm not here to scare you.  I'm here to help."\\\par

Lily stared at him, her confusion growing.  "Help?" she echoed, her voice filled with disbelief.  "Help with what?"\\\par

"With your food truck," Mr. Henderson said, his smile widening.  "I heard you've been struggling.  I know a thing or two about cooking, you see.  I used to be a chef, back in the day."
\\\par
Lily's eyebrows shot up in surprise.  "A chef?" she repeated, her skepticism battling with a sliver of curiosity.  "But...how can you help?  You're...you're a ghost."\\\par

Mr. Henderson chuckled again, his ghostly form shimmering faintly.  "I may be a ghost, but I still have a passion for food.  And I have a lot of experience.  Perhaps I can offer some advice, some insights, that might help you on your journey."\\\par

Lily's mind whirled.  The idea of a ghost offering business advice was absurd, preposterous even.  But there was something about Mr. Henderson, something in his gentle demeanor, his genuine concern, that made her hesitant to dismiss him outright.  She'd been struggling, feeling lost and alone, and the thought of a helping hand, even a spectral one, was strangely appealing.\\\par

"I...I don't know," Lily said, her voice barely a whisper.  "This is...this is a lot to take in."\\\par

"Take your time," Mr. Henderson said, his voice soothing.  "I understand.  It's not every day you meet a ghost who wants to give you business advice."\\\par

Lily couldn't help but smile at that, a flicker of humor breaking through her apprehension.  "No, it's not," she agreed, shaking her head.  "It's certainly not."
\\\par
Mr. Henderson's smile broadened.  "Perhaps we can have a cup of tea, get to know each other a little better?"\\\par

Lily hesitated, her mind a jumble of conflicting emotions.  But something inside her, a spark of hope, a desperate yearning for a solution, pushed her forward.\\\par

"Alright," she said, her voice a little stronger now.  "A cup of tea.  But I'm not promising anything."\\\par

Mr. Henderson's smile remained, his eyes twinkling with amusement.  "That's perfectly fine, Lily," he said.  "That's perfectly fine."\\\par

And so, in the dusty, sun-drenched living room of her childhood home, Lily found herself sitting across from a friendly ghost, sharing a cup of tea and a conversation that would change her life forever.  The road to success had always seemed long and winding, but with Mr. Henderson by her side, even if he was a little bit spectral, it felt like maybe, just maybe, she was finally on the right path.\\\par

Lily’s food truck business was booming.  Her once-empty parking lots were now filled with eager customers, the aroma of her signature dishes wafting through the air, drawing them in like moths to a flame.  The local newspapers had featured her, raving about her innovative dishes and her unique, whimsical approach to cooking.  She’d even won a prestigious award at a regional food truck competition, a testament to her talent and the power of Mr. Henderson’s guidance.\\\par

But with success came a new set of challenges.  The pressure to maintain her momentum, the constant demands of running a thriving business, the ever-present fear of losing her edge – these anxieties gnawed at her, leaving her feeling exhausted and overwhelmed.\\\par

One evening, after a particularly hectic day, Lily found herself back in the familiar, dusty living room of her childhood home.  Mr. Henderson, his spectral form shimmering in the soft glow of the lamplight, sat beside her, a cup of tea warming his translucent hand.\\\par

“You’re working too hard, Lily,” Mr. Henderson said, his voice a gentle reprimand.  “Remember what I told you about balance?  You need to take care of yourself, to make time for the things that matter.”\\\par

Lily sighed, her head resting in her hand.  “I know, Mr. Henderson,” she said, her voice weary.  “But it’s all so exciting, so overwhelming.  I’m afraid of letting it all slip away.”\\\par

Mr. Henderson smiled, his eyes twinkling with understanding.  “You’re not letting anything slip away, Lily,” he reassured her.  “You’ve built something special, something that will last.  But you need to remember that success is a journey, not a destination.  Enjoy the ride.”\\\par

Lily looked at him, her heart heavy.  “I don’t know, Mr. Henderson,” she said.  “I feel like I’m losing myself in all this.  I’m so focused on the food truck, on the business, that I’ve forgotten who I am.”\\\par

Mr. Henderson placed a gentle hand on her shoulder, his spectral touch sending a shiver down her spine.  “You’re never going to lose yourself, Lily,” he said, his voice a soft murmur.  “Your passion for food, your creativity, your kindness – these are all part of who you are.  Don’t let the business overshadow what truly makes you special.”\\\par

Lily’s eyes welled up, a wave of emotion washing over her.  She felt a deep sense of gratitude for Mr. Henderson, for his unwavering support, his gentle guidance, his ability to see through the noise and remind her of what truly mattered.\\\par

“What am I going to do, Mr. Henderson?” she asked, tears streaming down her cheeks.  “I feel so lost.”\\\par

Mr. Henderson smiled, his spectral form seeming to glow with a soft, warm light.  “You’re not lost, Lily,” he said, his voice a whisper of reassurance.  “You’re exactly where you’re supposed to be.  Just remember to breathe, to take a step back, to appreciate the journey.  And never, ever, forget who you are.”\\\par

Lily sat in silence, her eyes fixed on the flickering flames in the fireplace. Mr. Henderson’s words echoed in her mind, settling like a soothing balm on her troubled soul. She had been so focused on building her business, on chasing the elusive dream of success, that she had forgotten to take care of herself.\\\par

As the fire crackled and popped, a wave of exhaustion washed over her. The last few months had been a whirlwind of activity, a constant blur of cooking, serving, and strategizing. She had pushed herself to the limit, driven by a burning desire to prove herself, to make her grandmother proud. But in the process, she had lost sight of what truly mattered.\\\par

Lily stood up, stretching her stiff muscles. She needed to get out, away from the house, away from the weight of her anxieties. The desert night air, with its cool breeze and starry sky, might help to clear her head.\\\par

As she stepped out onto the porch, the silence of the night enveloped her. The only sounds were the distant howling of coyotes and the chirping of crickets. She took a deep breath, inhaling the scent of desert wildflowers and the cool, dry air. It was a reminder of the simple beauty that surrounded her, a beauty she had been too busy to notice.\\\par

Lily walked down the steps, her bare feet sinking into the soft sand. She began to walk, her footsteps echoing on the quiet street, the stars above twinkling like a million tiny diamonds. She walked until she reached the edge of town, where the desert stretched out before her, a vast, endless expanse of sand and sky.\\\par

She sat down on a weathered rock, her back against a gnarled mesquite tree. The stars twinkled above, casting a soft, ethereal glow on the desert landscape. Lily closed her eyes, trying to quiet the racing thoughts in her mind. She took another deep breath, allowing the cool night air to fill her lungs.\\\par

As she sat there, contemplating her life, she realized that Mr. Henderson was right. She had been so focused on the destination that she had forgotten to enjoy the journey. She had let the pressure of success cloud her judgment, making her forget the things that truly mattered.\\\par

Lily opened her eyes, a new sense of clarity flooding through her. She didn’t need to sacrifice her well-being for the sake of her business. She could have both – a thriving food truck and a happy, fulfilling life.\\\par

The desert night, with its vastness and silence, had given her the space she needed to reflect, to reconnect with herself. She had found a new perspective, a renewed sense of purpose. She would continue to work hard, to chase her dreams, but she would do so with a newfound balance, a commitment to taking care of herself.
\\\par
As the first rays of dawn began to paint the sky with shades of pink and orange, Lily stood up and started to walk back towards town. She had a lot of work to do, but she felt a sense of excitement, a renewed energy. She had learned a valuable lesson from Mr. Henderson, a lesson that would guide her on her path to success.\\\par

The following morning, Lily woke up with a renewed sense of purpose. The sun streamed through the dusty window, painting the room in a warm, golden light. The weight of anxiety that had been pressing down on her seemed to have lifted, replaced by a feeling of calm clarity.  She felt lighter, more grounded, as if the desert night had washed away the anxieties that had been clouding her mind.\\\par

She made herself a cup of strong coffee, the aroma filling the kitchen with a comforting scent. As she sipped her coffee, she thought about Mr. Henderson's words. He had reminded her of the importance of balance, of taking care of herself, of remembering who she truly was.\\\par

Lily took a deep breath, feeling a surge of determination. She would continue to chase her dream of a successful food truck, but she would do so with a newfound awareness of her own well-being. She wouldn't let the pressure of success overshadow her own happiness.\\\par

She decided to start by making some changes to her daily routine. She would take more breaks, schedule time for herself, and make sure to get enough sleep. She would also make an effort to connect with friends and family, to nurture the relationships that brought joy into her life.\\\par

Lily also decided to incorporate Mr. Henderson's advice into her business strategy. She realized that his insights, while sometimes outdated, were rooted in a deep understanding of food and a genuine love for culinary excellence. She would continue to experiment with new dishes, but she would also pay homage to the traditional recipes and techniques that Mr. Henderson had shared with her.\\\par

Lily spent the next few weeks putting her new plan into action. She hired a part-time assistant to help her with the workload, allowing her to take more breaks and spend more time with friends and family. She also started incorporating some of Mr. Henderson's recipes into her menu, adding a touch of old-world charm to her modern dishes.\\\par

The changes she made had a positive impact on both her personal life and her business. She felt less stressed and more fulfilled, and her customers seemed to appreciate the new additions to her menu.\\\par

One afternoon, as Lily was preparing a batch of Mr. Henderson's signature apple pie, she felt a familiar presence beside her.  She glanced over, her heart skipping a beat. Mr. Henderson stood there, his translucent form shimmering in the afternoon light. He smiled, his eyes twinkling with amusement.\\\par

"You've been doing well, Lily," he said, his voice a gentle murmur. "I'm proud of you."\\\par

Lily smiled back, a wave of warmth washing over her. "Thank you, Mr. Henderson," she said, her voice filled with gratitude. "I couldn't have done it without your guidance."\\\par

Mr. Henderson chuckled softly, his ghostly form seeming to glow with a soft, warm light. "You've found your own way, Lily," he said. "You've taken my advice and made it your own. You've created something special."\\\par

Lily nodded, her heart filled with a sense of accomplishment. She had learned so much from Mr. Henderson, not just about cooking and business, but about life itself. She had learned the importance of balance, of taking care of herself, of following her own path.\\\par

As the days turned into weeks and the weeks into months, Lily's food truck continued to thrive. She expanded her operation, hiring more staff and opening a second location. She became a respected figure in the local culinary scene, known for her innovative dishes and her commitment to quality.\\\par

But even as she achieved her dreams, Lily never forgot Mr. Henderson. She kept his recipes close at hand, and she often shared stories of him with her staff and customers. She knew that he was always with her, a ghostly mentor and friend, guiding her on her journey.\\\par

One evening, as Lily was closing up her food truck for the night, she noticed a faint shimmer in the air. She looked up, her heart filled with a mixture of excitement and sadness. Mr. Henderson stood there, his ghostly form shimmering with a gentle light.\\\par

"It's time for me to go, Lily," he said, his voice a whisper. "But I'll always be with you."\\\par

Lily's eyes welled up with tears. "Thank you, Mr. Henderson," she whispered. "Thank you for everything."\\\par

Mr. Henderson smiled, his eyes filled with a warmth that transcended the boundaries of life and death. "You've made me proud, Lily," he said. "Now go out there and make your own magic."\\\par

And with that, Mr. Henderson faded away, leaving Lily with a sense of peace and gratitude. She knew that she would never forget him, that his presence would always be with her, a reminder of the power of connection, of the importance of following your dreams, and of the magic that can happen when you open your heart to the unexpected.
    \end{spacing}
\end{tcolorbox}
    
\newpage
\begin{tcolorbox}[colback=white, colframe=my-green, coltitle=black, parskip=5mm, title=
    \textbf{Example Prompt}, breakable]
  \scriptsize
Write a science fiction story about someone who is a time traveler and has dedicated everything in their life towards a goal, and now wonders if it was worth it. The story should be between 850 and 900 words. The story should begin with the main character waking up on a frozen tundra. He looks for shelter from the cold. He sees a dead wooly mammoth and realizes he traveled back to the ice age. The character should find shelter, and a predator is outside his shelter at night. The ending should not be happy.
\end{tcolorbox}

\begin{tcolorbox}[colback=white, colframe=my-green, coltitle=black, parskip=5mm, title=
    \textbf{Human-written Story}, breakable]
  \scriptsize
        The cold wind pierced Robert’s exposed skin like hundreds of bee stings. Sprawled on his stomach on the solid, frozen ground, he felt chilled to his core. The light from the sun reflected off the endless stretch of ice. He struggled to open his eyes in the glare. \\\par
        Robert sat up and tried to regain his bearings. Although he was still foggy from the leap, he knew he must hasten to find shelter from the relentless cold.  \\\par
        “How could I be so ill prepared?” he berated himself aloud.
        The most he had ever traveled in time was forward seven months. He couldn’t control the date and time of his destination, but surely he had not time-jumped beyond civilization. Robert stood up and steadied himself on the ice. He looked around at the vast white wilderness and shook his head. This was the moment. The moment he acted out in his backyard as a child. The moment he had given up sleep to study entanglement and wormholes for. The moment for which he had sacrificed everything. He always imagined more pomp and circumstance and less uncertainty.  \\\par
        Robert tightened the hood of his sweatshirt over his head and pulled his hands into his sleeves. With no buildings in sight, he decided his best bet was to walk along the nearby riverbed to find a crevice or overhang that could provide shelter. The ground was hard and slick with a light dusting of snow. \\\par Sediment and rocks frozen on the surface helped provide some traction. A large dark mass appeared in the distance. As he neared the enormous object, the stench of rotten meat with the slightest note of sweetness grew stronger.  \\\par
        “No, no. It can’t be.” Robert audibly gasped.  \\\par
       Before him lay the ravaged carcass of a young wooly mammoth. Thoughts began to race and Robert grew dizzy. He fell to his knees before the massive tusks and began to dry heave. A combination of the putrid smell and the realization that he had actually traveled twenty thousand years into the past overwhelmed his mind and body. What had he gotten himself into?  \\\par
       Survival kicked in. He felt in his pocket for the hunting knife his grandfather gave him when he was a child. It had only ever been used to cut string or open packages. It had never been used on an actual animal. As rancid as the beast was, the fur would provide some protection from the biting cold. The skin was already a bit loose, and he cut through the ligaments to remove the pelt. He wrapped one piece around and felt immediate relief. He took another large piece to provide protection later.  \\\par
       Time travel took a huge toll on the body as it required a massive amount of energy. Pure adrenaline pushed Robert at this point. He must find shelter soon so that he could set up camp before sunset. There appeared to be a crevice in the rock beneath an overhang in the river bed. It was small but gave Robert enough room to stretch out. Not that he wanted to. All he wanted to do now was curl up under the pelt. He used large rocks to hold the second hide in place as a curtain in front of the opening.  \\\par
       He ventured only a few feet from his camp to gather branches from a fallen tree. A fire was critical, not only for warmth. It would hopefully keep away whatever had attacked that mammoth. He piled some wood near the crevice opening. He arranged some small sticks for tinder. Though he never got his fire making badge as a boy scout, he did use his glasses to burn ants when he was a child. Always playing God. What a fine mess that landed him in now.  \\\par
        He shivered from the cold but held his hand as steady as he could. The sun focused a beam of light on the tinder. A small trail of smoke appeared, and he blew on the glowing embers. Flames ignited, and Robert sobbed. He understood the exaltation the first human to discover fire must have felt.  \\\par
       The adrenaline was wearing off and reality began to set in. He sat in the shelter, hunched over with his arms wrapped around his legs. He stroked the wooly mammoth fur. When he had dreamed of time travel, he imagined watching the pyramids being built or leaping forward to see flying cars. Exhaustion finally overtook Robert, and he drifted off to sleep.  \\\par
       The fire had begun to smolder. Roberts' eyes popped open. Disoriented, he sat up, desperately searching for clues. Although he was completely surrounded by the blackness of night, the chill that ran through him reminded him of his reality. Outside the shelter, he heard a clicking on the ice. It stopped for a minute. Then it happened again.  \\\par
        A little louder and a little closer. Another pause. More tapping. The curtain moved and a sliver of moonlight shone through the opening. The light bounced off a long sharp fang, and Robert felt a hot breath on his arm. The creature let out a low growl. 
        Robert closed his eyes tightly and prepared for the final leap into the unknown.
\end{tcolorbox}
\newpage

 \begin{tcolorbox}[colback=white, colframe=gray!30!white, coltitle=black, parskip=5mm, title=
    \textbf{System output: \baselinezs}, breakable]
    \scriptsize
    \begin{spacing}{1}
The crunch of ice underfoot was the first thing I noticed. Then the biting wind, whipping across the frozen wasteland. I stumbled, the unfamiliar cold searing through my thin clothing. My breath formed white plumes in the air, instantly freezing on my face.  I was on a frozen tundra, a vast, white expanse stretching to the horizon.\\\par

It had been a long journey, a journey spanning millennia. My life, consumed by this singular goal, had finally brought me here. The Ice Age.\\\par

I scanned the landscape, searching for any sign of shelter. A lone, skeletal tree stood stark against the blinding white.  Then, a horrifying sight. A colossal carcass, its massive tusks jutting out from the snow, lay frozen in time. A wooly mammoth, its once-mighty form now a testament to the cruel indifference of nature.\\\par

I hurried towards it, the sight of the mammoth offering a glimmer of hope. Perhaps I could find some kind of shelter within its decaying remains. As I approached, the putrid stench hit me, a noxious mix of decay and death. I gagged, but pressed on. I had to find shelter before the sun dipped below the horizon and the cold truly bit.\\\par

Using my time-traveling device, I cut through the mammoth's tough hide, revealing a cavernous space within. The stench was overpowering, but it was better than freezing to death. I huddled inside, the decaying flesh a grim reminder of my isolation.\\\par

I felt a surge of exhaustion, a weariness that had built up over countless years of preparation, of relentless pursuit of this singular goal. Was it worth it? This desolate landscape, this chilling silence, this overwhelming sense of loneliness?\\\par

As darkness crawled across the sky, I huddled deeper into the carcass, the cold seeping through my clothes, a constant reminder of my vulnerability. The wind howled outside, a symphony of icy fury.\\\par

A guttural growl ripped through the silence. My heart leaped into my throat. A shadow flickered at the entrance of the makeshift shelter. My eyes, adjusting to the darkness, made out the silhouette of a massive beast, its amber eyes glowing in the moonlight. A saber-toothed tiger, its fangs glinting with deadly intent.\\\par

Panic surged through me. I had no weapons, no tools, no way to defend myself. I was trapped, a helpless prey in the jaws of an ancient predator.\\\par

I tried to back away, but the space was too small. The tiger, sensing my fear, pressed closer, its hot breath washing over me. It lowered its head, its fangs bared in a silent, deadly threat.\\\par

The smell of death, both from the mammoth and the tiger, filled the air. I closed my eyes, bracing for the inevitable. This was the end. The culmination of my life's work, the grand goal I had pursued with unwavering determination, had led to this - a gruesome death in the jaws of a prehistoric beast.\\\par

The tiger's growl, a low, guttural rumble, echoed in the cavern. I waited, frozen in fear, for the final, fatal blow. But it never came.\\\par

Instead, a different sound, a high-pitched whine, pierced the night. The tiger, startled, turned its head towards the sound. My heart pounded in my chest, a desperate hope flickering within me.\\\par

But then, silence. The tiger remained motionless, its amber eyes fixed on the darkness outside. The whine had stopped.\\\par

I held my breath, waiting for the attack to resume. But the silence held. Minutes stretched into an eternity, the tension in the air palpable. \\\par

Then, a faint, rhythmic thud, growing louder with each beat. The tiger’s head turned, its eyes widening in fear.\\\par

The thudding grew into a thunderous roar. Something massive, something much larger than the tiger, was approaching.\\\par

I couldn't see what it was, but the tiger's fear was contagious. It backed away, its growls now a whimper of fear, its tail tucked between its legs.\\\par

The earth shook as the creature drew closer, its heavy footsteps echoing in the night. The tiger, sensing its own insignificance in the face of this new threat, turned and fled, disappearing into the darkness.\\\par

I remained huddled in the carcass, a silent witness to the unfolding drama. I had escaped the immediate threat, but a new, even more terrifying one had arrived. I was alone, trapped in a hostile world, with no idea what awaited me in the darkness.\\\par

The chilling roar of the unknown creature echoed through the frozen wasteland, a sound that reverberated through my bones, a sound that told me my journey was far from over. And the question, "Was it worth it?" echoed even louder, a haunting reminder of the sacrifices I had made, the price I had paid for this moment. And the answer, a cold, stark truth, whispered back: "No." 

\end{spacing}

	\end{tcolorbox}
	
	\newpage 
	
	\begin{tcolorbox}[colback=white, colframe=gray!30!white, coltitle=black, parskip=5mm, title=
    \textbf{System output: \Agentsroomzs}, breakable]
    \scriptsize
    \begin{spacing}{1}
    
    The world was a canvas of white, stretching out endlessly in every direction. Elias, his breath forming frosty plumes in the frigid air, stumbled forward, his boots sinking into the powdery snow. The wind, a relentless beast, whipped at his face, stealing the warmth from his exposed skin.  He was alone, utterly and completely alone, on a frozen tundra that seemed to stretch back to the very dawn of time.\\\par

He had arrived. His journey, a life dedicated to a singular goal, had culminated in this moment.  He had traveled back in time, escaping the sterile confines of his lab and the relentless hum of his time machine, to stand on the frozen plains of the Ice Age. It was a dream realized, a goal he had pursued with an almost religious fervor, sacrificing everything else in his life to achieve it.\\\par

But now, the reality of the situation crashed down upon him with the force of a glacial avalanche. The stark beauty of the landscape, the vastness of the frozen world, was quickly eclipsed by the crushing weight of his isolation.  He was a lone figure, a speck of humanity lost in a world of ice and snow.  His heart pounded against his ribs, a frantic drumbeat in the silence of the tundra.\\\par

He had to find shelter. The cold was relentless, biting at his exposed skin, sapping his strength. He trudged forward, his eyes scanning the horizon for any sign of respite, any hint of protection from the elements.  The landscape was unforgiving, a vast expanse of white punctuated only by the occasional jagged rock protruding from the ice sheet.\\\par

Then, he saw it. A shallow cave, formed by a massive sheet of ice that had buckled and collapsed, offering a meager haven from the wind and the biting cold.  He stumbled towards it, his body shivering with exhaustion and the chilling cold.  He was desperate for warmth, for a moment of respite from the relentless assault of the elements.\\\par

As he entered the cave, the wind's icy fingers no longer reached him. The air inside was stale, heavy with the scent of frozen earth, but it was a welcome change from the biting wind that had been tearing at his face. The ice walls dripped and groaned, a constant reminder of the power of nature, the relentless force that ruled this frozen world.\\\par

He sank to the floor, his back against the cold, damp ice. He was exhausted, his body aching with the cold.  He closed his eyes, trying to shut out the harsh realities of his surroundings.  For a moment, he allowed himself to savor the victory, the culmination of his life's work. But the feeling was fleeting, quickly replaced by a growing sense of unease.\\\par

The isolation was suffocating. He was completely alone, cut off from the world he knew, the world he had left behind.  He had sacrificed everything for this moment, for this journey, and now, he was starting to question if it was truly worth it. The beauty he had sought, the world he had dreamed of experiencing, was now a source of dread.\\\par

He opened his eyes, his gaze falling on a sight that sent a shiver down his spine.  A woolly mammoth, its massive body frozen solid, lay sprawled in the snow just outside the entrance to his cave.  Its tusks were long and curved, its eyes wide and vacant, staring into the endless expanse of the frozen tundra. It was a stark reminder of the harsh reality of this world, a world where life was a constant struggle for survival.
\\\par

The mammoth's death was a chilling portent, a grim warning of the dangers that lurked in this unforgiving landscape.  He was not alone. He was prey.  And as the darkness settled over the frozen tundra, he could feel the presence of something else lurking in the shadows, a predator, watching, waiting.\\\par

The predator's presence was a constant in the back of his mind, a low hum of unease that vibrated through his bones. He couldn't shake the feeling of being watched, of being stalked. Every creak of the ice, every rustle of snow, sent a jolt of fear through him.  He tried to rationalize it, to tell himself it was just his imagination, the product of his exhaustion and the isolation gnawing at his sanity. But the primal instinct for survival, dormant for so long in his sheltered life, now roared to life, demanding attention.\\\par

He spent the next day huddled in his cave, the silence broken only by the drip-drip-drip of melting ice and the distant howl of the wind. He tried to focus on the beauty of his surroundings, on the vastness of the frozen landscape, on the intricate patterns etched into the ice by the relentless wind. But the beauty was tinged with fear, a constant reminder of the power of nature, the untamed wilderness he had sought out.\\\par

He tried to read, to escape into the pages of a worn copy of Darwin's "The Origin of Species," a book he had carried with him, a symbol of the knowledge he had sought. But the words blurred, the sentences lost their meaning, and the pages seemed to mock him with their irrelevant tales of a world he could no longer comprehend.\\\par

He tried to sleep, to escape the reality of his situation in the embrace of unconsciousness. But sleep eluded him. He lay there, his eyes wide open, listening to the silence, the silence that was broken only by the occasional snap of a twig or the rustle of snow.  He imagined the predator circling the cave, its keen eyes scanning the entrance, its powerful muscles coiled, ready to strike.\\\par

The fear was a physical sensation, a tightening in his chest, a clammy sweat on his skin. He was trapped, a lone figure in a world that had no room for him.  He was a visitor, an intruder in a world that had no place for his kind.\\\par

As darkness fell, the fear intensified. The cold seemed to bite deeper, the wind to howl louder, the silence to press in on him, suffocating him.  He huddled deeper into his cave, his back pressed against the ice, his hands clutching the tattered blanket he had brought with him. He was a small, insignificant creature, a speck of dust in the vastness of the frozen tundra.\\\par

He thought of his life, of the sacrifices he had made, of the years he had spent in pursuit of this moment. He thought of his family, of the friends he had abandoned, of the life he had left behind. He thought of the empty promises he had made to himself, the dreams he had chased, the goals he had achieved.\\\par

And now, he was here, alone, facing his own mortality. He was the product of his own ambition, a creature of his own making, and he was starting to realize that the price he had paid might have been too high.\\\par

The first crack in the ice came as a whisper, a barely audible groan that sent a tremor through the cave. Elias's heart leaped into his throat, his breath catching in his chest. He froze, every muscle in his body tense, listening. The silence stretched, a taut thread that threatened to snap. Then, a second crack, louder this time, a distinct splintering sound that echoed through the frozen chamber.\\\par

He scrambled back, his eyes searching the ice wall for the source of the sound.  A dark shape, a shadow against the pale light filtering through the entrance, began to take form. It was the wolf, its massive head wedged against the ice, its powerful jaws working at the frozen barrier.  The animal's eyes, two burning orbs of yellow fire, were fixed on him, reflecting the meager light of the moon that had begun to peek through the clouds.\\\par

A low growl rumbled from the predator's throat, a sound that vibrated through the cave, a sound that spoke of primal hunger, of relentless instinct. Elias felt a wave of nausea rise within him, the taste of fear bitter on his tongue. He was trapped, a small, insignificant creature facing a predator that embodied the raw power of nature.
\\\par

He was alone, completely alone, with no hope of escape, no chance of rescue.  The years of sacrifice, the relentless pursuit of his goal, had brought him to this moment, a moment of absolute and utter despair.  He had traded his life for this, for this frozen wasteland, for this terrifying confrontation with his own mortality.\\\par

His gaze fell on the mammoth's frozen body, its lifeless eyes staring into the vastness of the tundra. It was a mirror, reflecting his own fate, a testament to the harsh realities of this world.  He had sought out this beauty, this unforgiving landscape, and now, he was about to become a part of it, a frozen relic in a world that had no place for him.\\\par

His fingers tightened around the tattered blanket, his last remaining comfort, his last connection to the world he had left behind.  He was a prisoner in his own creation, trapped in a cage of his own making.  He had sacrificed everything for this moment, and now, he was faced with the ultimate consequence of his ambition, a confrontation with the predator that lurked within himself.\\\par

The wolf's relentless assault on the ice wall was a symphony of destruction, each crack and groan a hammer blow to Elias's already shattered spirit. He pressed himself further into the back of the cave, his heart a frantic drum against his ribs.  He was a cornered animal, facing a predator that seemed to embody the very essence of the Ice Age.\\\par

His gaze swept across the cave, landing on the mammoth's frozen form. It lay there, a silent testament to the brutal reality of this world, a world where life was a constant struggle for survival.  The mammoth's fate was a chilling reflection of his own, a stark reminder that he was not immune to the laws of nature.\\\par

He tried to summon the courage to fight back, to defend himself against the relentless assault, but the fear was a heavy weight on his chest, suffocating his will.  He was trapped, a small, insignificant creature facing the raw power of nature.\\\par

The wolf's growls intensified, a chorus of primal hunger that echoed through the cave. Each sound was a blow to his spirit, a reminder of the futility of his struggle. He closed his eyes, trying to shut out the sounds, the smells, the overwhelming sense of dread that permeated the air.\\\par

He thought of his life, of the years he had spent in pursuit of this moment, of the sacrifices he had made.  He thought of the dreams he had chased, the goals he had achieved, and the emptiness that now filled his soul.  He had traded his life for this, for this moment of terror, for this confrontation with the predator that lurked within himself.\\\par

The wolf's teeth ripped through the ice, sending a shower of shards flying.  Elias flinched, his body trembling with fear.  He could feel the cold air on his face, the smell of the predator's breath.  He was on the brink, a precipice of despair from which there was no escape.\\\par

He wished he could go back, to undo the years of obsession, to reclaim the life he had abandoned.  He wished he could  find solace in the familiar comforts of his own time, in the warmth of human connection, in the simple joys of a life lived in balance.  But those wishes were just echoes now, fleeting whispers in the deafening roar of his own despair.\\\par

The wolf's head finally broke through the ice, a jagged maw of teeth and fury emerging into the meager light of the cave. Elias, his body frozen by fear, could only stare as the creature's eyes, two burning embers of predatory hunger, locked onto his. The air within the cave seemed to crackle with anticipation, a palpable tension that hung heavy in the frigid air.\\\par

A final, guttural growl ripped from the wolf's throat, a sound that seemed to echo the shattering of Elias's own spirit. He knew, in that moment of terrifying clarity, that there was no escape. His journey, his life's obsession, had led him to this: a final, brutal confrontation with the harsh realities of the world he had sought to witness.\\\par

The wolf lunged, a blur of muscle and teeth, its powerful jaws snapping shut just inches from Elias's face. He screamed, a primal sound of terror that echoed through the cave, a desperate plea for a mercy that would never come. The wolf's teeth sank into his arm, a searing pain that ripped through him, a reminder that his journey had been nothing but a fool's errand.\\\par

He felt the wolf's weight on him, the creature's hot breath against his skin, the stench of raw, predatory instinct. He struggled, but it was a futile effort, a desperate dance against the inevitable. He could feel the life draining from him, the warmth of his blood mingling with the cold of the ice.\\\par

In his fading vision, he saw the mammoth's frozen form, a chilling reminder of the finality of his own fate. He had sought out beauty, knowledge, a glimpse into the past, but all he had found was an unforgiving wilderness, a world of constant struggle and a predator's hunger.\\\par

His last thought, as darkness claimed him, was a whisper of regret, a silent admission that his journey had been a tragic mistake, a pursuit of knowledge that had cost him everything. The predator's victory was complete, its hunger satisfied, its instinct fulfilled. Elias, the obsessed time traveler, was just another frozen relic in the vast, unforgiving landscape of the Ice Age, a testament to the dangers of ambition and the emptiness of a life dedicated solely to a singular goal.\\\par
    \end{spacing}
    \end{tcolorbox}
	
\section{Specialized Agents' Prompt Templates}
\label{appendix:agent_prompts}

Each specialized LLM-based agent has a specific prompt template that formats the scratchpad into a suitable prompt for the LLM call.

\subsection{Scratchpad}

Concretely, the scratchpad is typically formatted as follows: 

\begin{tcolorbox}[colback=white, colframe=my-purple, coltitle=black, parskip=5mm, title=
\textbf{\textsc{[Scratchpad]} Format}, breakable, halign=flush left]
[Creative Writing Task]
$<$the original writing prompt$>$
\bigskip

[Central Conflict]
$<$the output of the conflict agent$>$
\\
\medskip 

[Character Descriptions]
$<$the output of the character agent$>$\\
\medskip

[Setting]
$<$the output of the setting agent$>$
\\
\medskip
[Key Plot Points]
$<$the output of the plot agent$>$
\\
\medskip
[Exposition]
$<$the output of the exposition agent$>$\\
\medskip

[Rising Action]
$<$the output of the rising action agent$>$\\
\medskip

[Climax]
$<$the output of the climax agent$>$
\\
\medskip
[Falling Action]
$<$the output of the falling action agent$>$
\\
\medskip
[Resolution]
$<$the output of the resolution agent$>$
\end{tcolorbox}

The number and order of items in the scratchpad is of course a function of which agents have been called so far.

\subsection{Planning Agents}
\label{appendix:planning_agents_prompts}

\begin{tcolorbox}[colback=white, colframe=my-blue, coltitle=black, parskip=5mm, title=
\textbf{\textsc{[Conflict]} Agent Prompt}, breakable, halign=flush left]
Given 
$<$identifiers found in the scratchpad$>$,
         describe the central conflict in detail (more than 5 sentences). The description should answer the following questions: 
         \begin{itemize}
        \item[{$\star$}] What's the protagonist's main goal in this story?
        \item[{$\star$}] Why do they want it? 
        \item[{$\star$}] What's stopping them from achieving it?
        \end{itemize}
$<$scratchpad$>$
        \end{tcolorbox}
        
\begin{tcolorbox}[colback=white, colframe=my-blue, coltitle=black, parskip=5mm, title=
\textbf{\textsc{[Character]} Agent Prompt}, breakable]
Given 
$<$identifiers found in the scratchpad$>$,
        describe the characters in detailed bullet points (more than 5
         sentences for each character). The description should answer the
        following questions:
        \begin{itemize}
            \item[{$\star$}] What do the characters sound like? Are they
         talkative or quiet? What kind of slang do they use? What is their
         sense of humor like?
         \item[{$\star$}] What do they look like? Do they have any
         defining gestures? What's the first thing people notice about
         them?
         \item[{$\star$}] What are their motivations and internal characteristics?
         What are their flaws? What are their values? What are they afraid of?
         How will they change and grow over the course of this story?
         \end{itemize}
$<$scratchpad$>$
\end{tcolorbox}

\begin{tcolorbox}[colback=white, colframe=my-blue, coltitle=black, parskip=5mm, title=
\textbf{\textsc{[Setting]} Agent Prompt}, breakable]
Given  
$<$identifiers found in the scratchpad$>$, 
         describe the setting in detail (more than 5 sentences). The
         description should answer the following questions:
         
         \begin{itemize}
             \item[{$\star$}] Where does the
         story take place? Is it set in a fictional world, or is it simply set
         in someone's backyard?
        \item[{$\star$}] When does the story take place? What decade
         is it set in? How much time elapses over the course of the story?
         \end{itemize}
$<$scratchpad$>$
\end{tcolorbox}

\begin{tcolorbox}[colback=white, colframe=my-blue, coltitle=black, parskip=5mm, title=
\textbf{\textsc{[Plot]} Agent Prompt}, breakable, halign=flush left]
Given 
$<$identifiers found in the scratchpad$>$,
         describe the key plot points in detailed bullet points.
\medskip
\\
$<$scratchpad$>$
\end{tcolorbox}

The $<$identifiers found in the scratchpad$>$ are extracted from the scratchpad and formatted to fit the prompt. For instance, for a scratchpad that contains the original prompt, the {\sc [conflict]} and {\sc [character]} agents' contributions, the resulting $<$identifiers found in the scratchpad$>$ would be: “a Creative Writing Task, the Central Conflict, and the Character Descriptions".

\subsection{Writing Agents}

\begin{tcolorbox}[colback=white, colframe=my-red, coltitle=black, parskip=5mm, title=
\textbf{\textsc{[$<$Section$>$]} Agent Prompt}, breakable, halign=flush left]
Given 
$<$identifiers found in the scratchpad$>$,
continue the story by writing the $<$section$>$ part.
\medskip

$<$If previous sections have been written, include the following in the prompt:$>$

\medskip
Begin your portion of the story in a way that naturally flows from the previous ending. Match the writing style, vocabulary, and overall mood of the existing text. Do not re-explain details or events that have already been described.
\medskip 

$<$If this is not the meant to be the last section, include the following in the prompt:$>$

\medskip
Focus only on the $<$section$>$ part of the story. Do not write about the following parts of the story. Do not end the story.

\medskip
$<$scratchpad$>$
\end{tcolorbox}

In these writing agents' prompt templates:

\begin{itemize}
    \item $<$section$>$ is one of “Exposition", “Rising Action", “Climax", “Falling Action", or “Resolution", 
    \item $<$identifiers found in the scratchpad$>$ are extracted from the scratchpad and formatted to fit the prompt. For these writing agents they are formatted as follows: 
“a Creative Writing Task, the Content Plan (Central Conflict, Character Descriptions,  Setting, Key Plot Points), and the Previous Parts of the Story (Exposition, Rising Action, Climax)", where the enumerated elements correspond to what is in the scratchpad.
\end{itemize}

In the specific case of the \Agentsroom$\;${\sc [planning]} variant, with only the planning agents, we still need a single writing agent to finalize the story, which we denote as the {\sc [finalizer]}. This {\sc [finalizer]} agent uses the following prompt template: 

\begin{tcolorbox}[colback=white, colframe=my-red, coltitle=black, parskip=5mm, title=
\textbf{\textsc{[Finalizer]} Agent Prompt}, breakable, halign=flush left]
Given 
$<$identifiers found in the scratchpad$>$,
         write a story using the information below.
         
\medskip
$<$scratchpad$>$
\end{tcolorbox}

\section{Prompt Templates for Synthetic Data Generation}
\label{appendix:synthetic_data_generation}

For the planning agents, we use the same prompt templates as in \Cref{appendix:planning_agents_prompts} to generate the synthetic training data, except in this case, we provide the gold standard data in the scratchpad. As a consequence, the scratchpad is formatted as follows: 

\begin{tcolorbox}[colback=white, colframe=my-purple, coltitle=black, parskip=5mm, title=
\textbf{\textsc{[Scratchpad]} Format}, breakable, halign=flush left]
[Creative Writing Task]
$<$the original writing prompt$>$
\bigskip

[User-Written Response]
$<$the gold output$>$
\end{tcolorbox}

The $<$identifiers found in the scratchpad$>$ are formatted as “a Creative Writing Task and a User-Written Response".

For the writing agents, we use the following prompt template to split to gold standard stories into distinct sections: 

\begin{tcolorbox}[colback=white, colframe=my-red, coltitle=black, parskip=5mm, title=
\textbf{\textsc{[Writing]} Synthetic Data Generation}, breakable, halign=flush left]
Split the following story into sections:
\begin{itemize}
\item[{$\star$}] [Exposition]: The exposition gives the reader the background info they need to jump right into the story's world. This is often found towards the beginning of the story.

\item[{$\star$}] [Rising Action]: The rising action is the moments in the story that lead up to the climax — choices the main characters have made and the events happening that are at odds with the characters' goals. This is where the story builds and the reader begins to invest in the characters.

\item[{$\star$}] [Climax]: The climax is the primary turning point and what the story has been building towards.

\item[{$\star$}] [Falling Action]: The falling action is the period of time in a story that follows the climax and leads to the resolution. It can be used to clarify the events of the climax, ease any built-up tension, or wrap up loose ends.

\item[{$\star$}] [Resolution]: This is the end of the story. It answers the remaining unanswered questions in the plot. The resolution is also the time to show the next step in the characters' lives.
\end{itemize}

For each section, give the section header (marked as [Exposition], [Rising Action], [Climax], [Falling Action], and [Resolution]) followed by the first sentence of that section, copied exactly from the story.
\bigskip

[User-Written Response]
$<$the gold output$>$
\end{tcolorbox}
\newpage

\section{Human Evaluation Instructions}
\label{appendix:instructions} 

For this task, you will be presented with a writing prompt and two  short stories corresponding to this prompt. Your task is to compare the quality of the two stories across several dimensions. This is a judgment task rather than an annotation task. As such, you should use your own judgment when you assign ratings, calibrated by the rubrics we provide.

This rating task consists of three steps: (1)~Compare the quality of the two stories across four dimensions. (2)~Rate which story you preferred. (3)~(optional)  Leave comment / feedback on the stories.
In the following we provide detailed instructions for each step: 

\subsection{Rate the Quality of the Story}
Your task is to compare the quality of two stories along four different dimensions (plot, creativity, development, language use), as described in the Rubric table below. 

While the dimensions may have overlap and work in interdependent ways, they are intended to capture distinct aspects of what makes a good story. Therefore, a story may score highly in one dimension and poorly in another. Furthermore, the features (marked as a, b, c) that make up a dimension may be thought of as cumulative. For example, a story may have strong characters but suffer from an underdeveloped setting. 

The Rubric table is intended to help you calibrate your judgment so that you can roughly determine when a story is very good or even excellent along a particular dimension because it exhibits all of the features of that dimension. Conversely, if a story fails to exhibit most or all of the features of a dimension, then you may score the story as being poor or very poor along that dimension. The features are meant to be illustrative but not exhaustive; you may determine that a story should score poorly or well due to the absence or presence of additional features for a given dimension based on your judgment.

Another important thing to note about the features that make up the dimensions we’re asking you to rate is that they describe conventions that may be followed or flouted; a story may contain intentional plot devices like non-linear timelines, discontinuity, and other stylistic choices to create effects. As with other features, these elements of a story should inform your judgment on their own merit (so that they only negatively impact your rating if they are ineffective or confusing and positively impact your rating if they are used well to make the story more interesting and unique).

We use a 3-point comparative rating scale for each of the dimensions. The rating scale can be thought of as described below:

\medskip
\begin{tabular}{|l|l|} \hline
\multicolumn{1}{|l|}{\textbf{Rating} } & \\\hline
A is better &
Response A is better than Response B in that dimension.\\
About the same & 
Both responses are about the same in that dimension. \\
B is better & 
Response B is better than Response A in that dimension.\\ \hline
\end{tabular}
\medskip

The focus of this rubric is the quality of the writing, and not how well the stories follow the writing prompt. In particular, when rating with this rubric, we encourage you not to focus on the number of words mentioned in the writing prompts, but rather on the features described in the  table below. 
\medskip

\begin{tabular}{|p{3cm}|@{~~}l@{~~}p{10cm}|} \hline
\textbf{Dimension} &  \multicolumn{2}{c}{\textbf{Features}}\\\hline
Plot & a. & The story has a recognizable structure, e.g. with a connected beginning, middle, and end.
\\
& b. & The story exhibits events and turns that move the plot forward. \\
& c. & The story does not have logical or conceptual inconsistencies. Surprising or disruptive elements are intentional, e.g., they serve the story and do not feel jarring, odd, or out of place.\\\hline
Creativity of Ideas, Themes, and Topics & 
a. & Engaging characters, themes, and imagery. The ideas do not feel generic or bland. \\
& b. & Avoidance of overly cliched characters and storylines, unintentional tropes, and stereotypes. When used, tropes and cliches serve a purpose (e.g. comedic effect, twist on a common trope etc). 
\\
& c. &The story includes original elements that were not explicitly mentioned in the prompt.\\\hline
Development & 
a. &Characters and settings are introduced and contextualized with relevant details that allow the reader to understand their place in the story.\\
& b. & Appropriate levels of detail and complexity are provided to lend the story a feeling of realness and believability.\\ 
& & \\
&  & \emph{Reminder: The features that make up a dimension may be thought of as cumulative. A story with a well-developed character, but in a lackluster setting (or vice-versa) would score lower in Development than a story that does well on both aspects.} \\\hline
Language Use & 
a. & The language used feels varied and rich: Variance of sentence structure, verbiage, and vocabulary. 
\\
& b. & The story exhibits rhetorical, linguistic and literary devices (e.g. ambiguity, alliteration, etc) to create interesting effects\\
& c. & The story avoids bland or repetitive phrases (unless used intentionally to create a narrative, thematic, or linguistic effect). 
\\\hline
\end{tabular}

\medskip

We provided examples rated along these rubrics. While the examples include explanations, these are there as an aid, and you are not requested to provide explanations for your ratings. 

\subsection{Which story do you prefer?}
Do you find the story interesting, engaging, funny, or emotionally-rich? In addition to getting your judgments of the dimensions, we would also like to know whether you enjoyed reading the story. Similar to the dimensions, we will ask you to score which story you prefer: \begin{itemize}
\item A is better
\item About the same
\item B is better
\end{itemize}

When rating, do not hesitate to be very critical.

\subsection{Optional: Leave comments or feedback on the stories}
Thank you for completing the ratings! If you have any additional comments or feedback you would like to provide on the story, feel free to add them in the “comments” section.

\section{Prompt Template for the LLM Evaluator}
\label{appendix:sxs_template}

The following prompt template is used by the LLM to evaluate two system outputs \textit{side-by-side}  (we replace $<$story a$>$ and $<$story b$>$ with the two stories being evaluated):

\begin{tcolorbox}[colback=white, colframe=my-yellow, coltitle=black, parskip=5mm, title=
\textbf{\textsc{[LLM Evaluator]} Prompt Template}, breakable]
You will conduct a side-by-side evaluation. You will be given two
system-generated stories. Your task is to compare the two stories and
determine which one is better based on the following dimensions:\\
\begin{itemize}
\item \emph{Plot:} The story should have a recognizable structure, e.g., with a
connected beginning, middle, and end. The story should exhibit events and
turns that move the plot forward. The story should not have logical or
conceptual inconsistencies. Surprising or disruptive elements should be
intentional, e.g., they serve the story and do not feel jarring, odd, or
out of place.
\item \emph{Creativity:} There should be engaging characters, themes, and imagery. The
ideas should not feel generic or bland. There should be avoidance of
overly cliched characters and storylines, unintentional tropes, and
stereotypes. When used, tropes and cliches should serve a purpose (e.g.,
comedic effect, twist on a common trope etc). The story should include
original elements that were not explicitly mentioned in the prompt.
\item \emph{Development:} Characters and settings should be introduced and
contextualized with relevant details that allow the reader to understand
their place in the story. Appropriate levels of detail and complexity
should be provided to lend the story a feeling of realness and
believability.
\item \emph{Language Use:} The language used should feel varied and rich: Variance of
sentence structure, verbiage, and vocabulary. The story should exhibit
rhetorical, linguistic and literary devices (e.g., ambiguity,
alliteration, etc) to create interesting effects. The story should avoid
bland or repetitive phrases (unless used intentionally to create a
narrative, thematic, or linguistic effect).
\end{itemize}
\bigskip 

Provide a detailed assessment of the two stories in terms of these four
dimensions. Conclude your assessment with scores for each dimension using
the template below. Do not add any emphasis, such as bold and italics, on
your assessment.\\
$[$Assessment Ending Template$]$\\
\bigskip

\begin{center}
Based on my assessment, the better story for each dimension is:\\

\medskip

\begin{tabular}{ll}
Plot: & [A or B or Same]\\
Creativity: & [A or B or Same]\\
Development: & [A or B or Same]\\
Language Use: & [A or B or Same]\\
Overall: & [A or B or Same]\\
\end{tabular}
\bigskip

$[$Story A$]$\\
$<$story a$>$\\
\quad \\
$[$Story B$]$\\
$<$story b$>$\\
\quad \\
$[$Assessment$]$
\end{center}
\end{tcolorbox}

Below we show example output from the LLM evaluator from which we extract preference scores. In this case,  the evaluator preferred story A on all dimensions. 

\begin{tcolorbox}[colback=white, colframe=my-yellow, coltitle=black, parskip=5mm, title=
\textbf{\textsc{[LLM Evaluator]} Example Output}, breakable]
Story A has a more defined plot structure with a clear beginning, rising action, climax, and resolution. The summoning of Horus, his reaction to the modern world, and his eventual acceptance create a compelling narrative arc. Story B, while touching upon interesting themes, lacks a clear climax and resolution. The story feels incomplete, leaving the reader with unanswered questions about Horus's fate and Ash's journey.\\
\quad \\
Story A exhibits more creativity in its portrayal of Horus and his interactions with Amelia. The falcon god's bewilderment, anger, and eventual acceptance are depicted with nuance and depth. The story also introduces original elements, such as the restoration of the canopic jar, which adds a layer of symbolism and reinforces the theme of reconciliation. Story B relies on more generic depictions of a god's reaction to the modern world, lacking the originality and depth seen in Story A.\\
\quad \\
Both stories provide adequate development of their main characters, Amelia and Ash, respectively. However, Story A excels in its development of Horus, providing a more nuanced and believable portrayal of a god grappling with the realities of a changed world. The story delves into his emotions, his struggles, and his eventual acceptance, making him a more compelling and relatable character. Story B's portrayal of Horus, while touching upon his disillusionment, lacks the same level of depth and complexity.\\
\quad \\
Story A utilizes a richer and more varied language, employing vivid imagery, sensory details, and a wider range of vocabulary. The story effectively uses rhetorical devices, such as repetition and contrast, to emphasize the differences between the ancient and modern worlds. Story B, while adequately written, lacks the same level of linguistic richness and sophistication.

\medskip 
\begin{center}
Based on my assessment, the better story for each dimension is:\\
\medskip
\begin{tabular}{ll}
Plot: & A\\
Creativity: & A\\
Development: & A\\
Language Use: &A\\
Overall: &A
\end{tabular}
\end{center}
\end{tcolorbox}

\section{Additional Results}
\label{sec:additional-results}

While we consider human-based evaluation our primary means of
evaluation, the LLM evaluator helps us assess overall system-level
trends. We report pairwise win rate (proportion of examples on which
our \Agentsroom~{\it plan + write} variant performed better than
comparison systems according to our LLM-based
evaluator. Table~\ref{tab:additional-results} complements~\Cref{fig:human_autorater_evals}a.

  \begin{table}[h!]
\centering
    \begin{tabular}{lccccc} \toprule
\agentsroomzs~plan + write vs  & overall &	plot &	creativ. &	develop. &
language \\ \midrule
 \baselinezs\ plan   &	74.55  &	63.64  &	75.47  &	75.93 & 	81.13\\
\baselinezs\ reflect  &	67.27  &	63.64  &	67.92  &	68.52  &	69.23\\
\baselinezs\ decompose & 	89.09  &	80.00  &	87.04  &	90.91  &	90.91\\
\textsc{2Stage} decompose 	 &66.67  &
59.26 & 	64.15  &	66.67 & 	67.92\\ \bottomrule
    \end{tabular}
    \caption{\label{tab:additional-results} Proportion of times
      LLM-based evaluator preferred \agentsroomzs\ to comparison system
      across overall, and across the dimensions of plot, creatitivy,
      development, and language use.}
  \end{table}

The majority of our experiments were conducted using a Gemini 1.5
Flash backbone model. This choice was dictated by the nature of the
creative writing task which is challenging to accomplish with models
that do not have a long enough context window and adequate writing
quality. Most recent work on storytelling using a single model resorts
to large, proprietary models such as GPT
\citep{yang-etal-2023-doc,yang-etal-2022-re3}, or Claude
  \citep{chakrabarty:ea:2024a}. This is also the case for multi-agent
  systems targeting writing which seem to be exclusively relying on
  GPT-4
  \citep{ijcai2024p3,bai2024longwriterunleashing10000word}. Nevertheless,
using Gemma2-9B-it  \citep{Riviere2024Gemma2I} as a backbone model we
compare \Agentsroom\ and \baseline\ systems in the zero-shot setting,
using the LLM-based evaluator.

\begin{table}[h!]
\centering
    \begin{tabular}{lcccccc} \toprule
  \Agentsroomzs\  plan+write vs.                         & overall &	plot &	creativ.&
      develop. &	language  \\\midrule
 \baselinezs\ & 	80.00& 	67.27& 	84.62&
83.33 &	77.78 \\ \bottomrule
    \end{tabular}
    \caption{\label{tab:gemma} Proportion of times LLM-based evaluator
       preferred \Agentsroomzs\  to  \baselinezs\ overall, and across the
      dimensions of plot, creativity, development, and language use.}
\end{table}

As can be seen in Table~\ref{tab:gemma}, even with the  smaller scale
Gemma2-9B-it model, \Agentsroom\ greatly
outperforms the end-to-end baseline across all dimensions of
evaluation. 

Finally, although we did not elicit feedback on individual story dimensions, we did ask participants to comment on the quality  of the stories produced by our systems, and possibly on aspects of story quality our instructions did not cover (see Section~\ref{appendix:instructions}). We show some of this feedback below. 

\begin{tcolorbox}[colback=white, colframe=my-yellow, coltitle=black, parskip=5mm, title=
\textbf{Participant Feedback}, breakable]
"The task was interesting, but over time, I found the language redundant. There seemed to be a go-to vocabulary list utilized in the majority of the stories, phrases used time and again, making the output somewhat predictable." \tcbline

"It was interesting to see what kind of fictional narrative the model would generate. Most of the stories seemed to be written at a seventh grade level. The stories didn't stray too far from the input and for the most part were grammatically correct. There were at times, instances of repetitiveness, including entire paragraphs, that made me wonder what the model was doing."\tcbline

"The stories showed some promise, but often fell into the same pitfalls of loops or sudden tone discordance…"
\end{tcolorbox}

\end{document}